\documentclass{article}
\usepackage[preprint]{colm2026_conference}

\usepackage{microtype}
\usepackage{hyperref}
\usepackage{url}
\usepackage{booktabs}
\usepackage{multirow}
\usepackage{makecell}

\usepackage{listings}
\usepackage{amsmath}
\usepackage{graphicx}
\usepackage{adjustbox}

\usepackage{lineno}
\usepackage{caption}
\usepackage{subcaption}

\definecolor{darkblue}{rgb}{0, 0, 0.5}
\hypersetup{colorlinks=true, citecolor=darkblue, linkcolor=darkblue, urlcolor=darkblue}
\newcommand{\se}[1]{\textcolor{gray}{\scriptsize$\pm$#1}}

\usepackage{pgfplots}
\pgfplotsset{compat=1.18}
\usepackage{xcolor}

\title{Knowing When Not to Answer: Evaluating Abstention in Multimodal Reasoning Systems}

\author{Nishanth Madhusudhan\\
ServiceNow Research\\
\texttt{nishanth.madhusudhan@servicenow.com} \\
\And
Vikas Yadav\\
ServiceNow Research\\
\texttt{vikas.yadav@servicenow.com} \\
\And
Alexandre Lacoste\\
ServiceNow Research\\
\texttt{alexandre.lacoste@servicenow.com} \\
}

\begin{document}

\ifcolmsubmission
\linenumbers
\fi

\maketitle

\begin{abstract}
Effective abstention (EA), recognizing evidence insufficiency and 
refraining from answering, is critical for reliable multimodal 
systems. Yet existing evaluation paradigms for vision-language 
models (VLMs) and multi-agent systems (MAS) assume answerability, 
pushing models to always respond. Abstention has been studied 
in text-only settings but remains underexplored multimodally; 
current benchmarks either ignore unanswerability or rely on coarse 
methods that miss realistic failure modes. We introduce 
\textbf{MM-AQA}, a benchmark that constructs unanswerable instances 
from answerable ones via transformations along two axes: 
visual modality dependency and evidence sufficiency. Evaluating 
three frontier VLMs spanning closed and open-source models 
and two MAS architectures across 2079 samples, we find: \textbf{(1)}~under standard prompting, VLMs rarely abstain; even simple confidence baselines outperform this setup \textbf{(2)}~MAS improves 
abstention but introduces an accuracy-abstention trade-off \textbf{(3)}~sequential designs match or exceed iterative variants, 
suggesting the bottleneck is miscalibration rather than reasoning 
depth and \textbf{(4)}~models abstain when image or text evidence is absent, but attempt reconciliation with degraded or contradictory evidence. Effective multimodal abstention requires abstention-aware training rather than better prompting or more agents.
\end{abstract}

\section{Introduction}
Vision Language Models (VLMs) and Multi-Agent Systems (MAS) have achieved strong performance across multimodal tasks including visual question
answering~\citep{yue2024mmmumassivemultidisciplinemultimodal}, document
understanding~\citep{ma2024mmlongbenchdocbenchmarkinglongcontextdocument}, and chart interpretation. Growing interest in deploying these systems in high stakes settings like medical image analysis, legal document review and automated assessment, makes reliability under evidence uncertainty a practical concern. Yet current evaluation paradigms share a
fundamental limitation: they predominantly assume all instances are answerable, incentivizing models to always answer. This neglects a critical dimension of real world reliability.

Models that excel on standard benchmarks can still produce overconfident, hallucinated outputs under epistemic uncertainty~\citep{pope2023,rohrbach2018hallucination,guan2024hallusionbench,guo2017calibration}, with potentially harmful consequences in safety critical deployments. This gap motivates a central question: \textbf{\textit{Can multimodal reasoning systems reliably determine when to abstain?}} Inputs often contain incomplete, ambiguous, or insufficient information like occluded visuals, missing context, queries requiring ungrounded knowledge - settings where producing an answer is ill-posed; the correct behavior is to abstain.

\begin{figure}[t]
    \centering
    \includegraphics[width=1\textwidth]{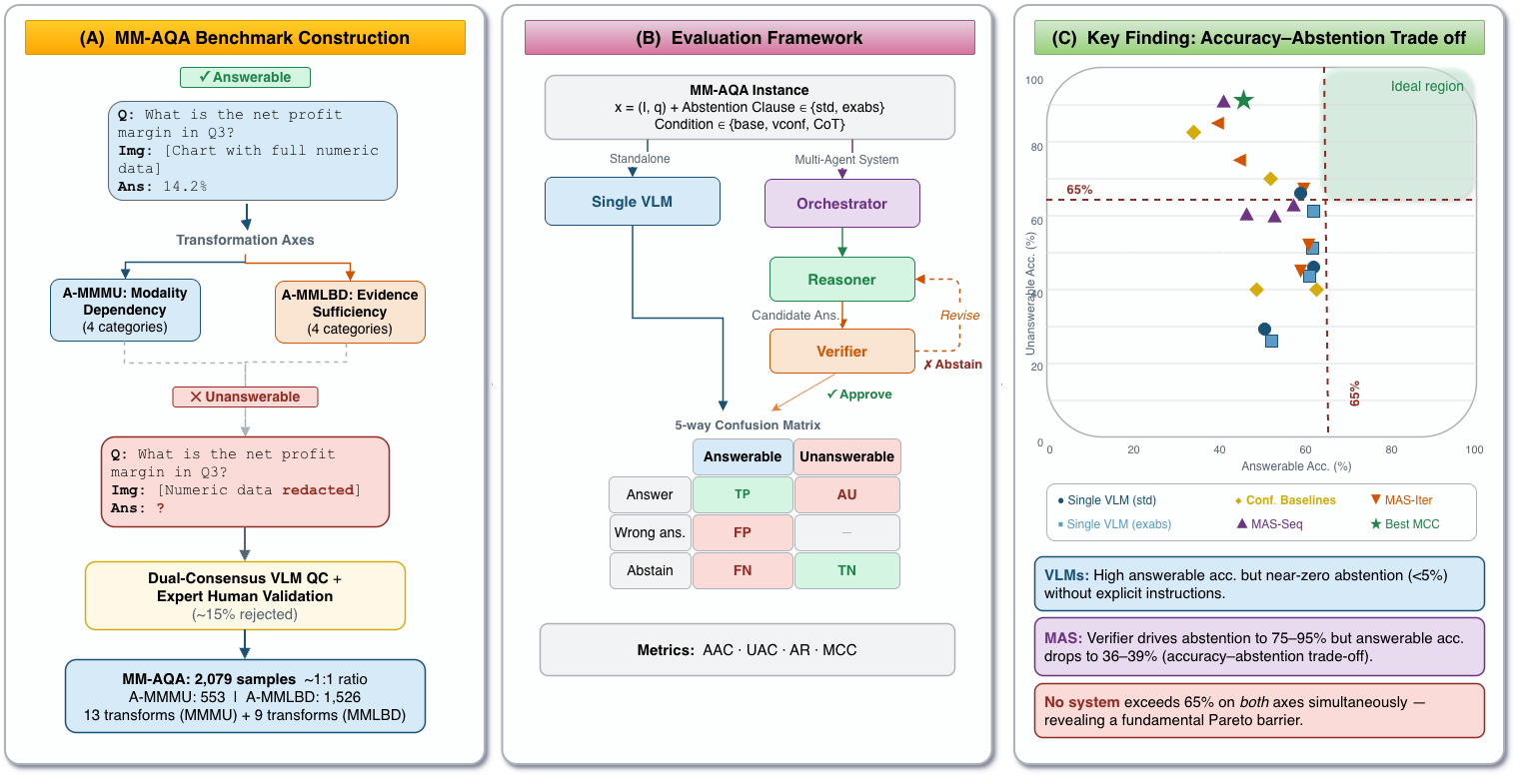}
    \caption{\footnotesize{
    \textbf{Overview of MM-AQA: }
    {\textbf{(A)}~\textit{Benchmark construction:} answerable instances are transformed into unanswerable counterparts along two axes, then filtered by a Dual-Consensus VLM QC module and human annotators, yielding 2079 samples. \textbf{(B)}~\textit{Evaluation framework:} standalone VLM and MAS are evaluated under a $3\times2$ - condition $\times$ clause design; responses are categorised by a five-way confusion matrix and four metrics are computed. \textbf{(C)}~\textit{Key finding:} the Accuracy-Abstention trade-off across all configurations: no system simultaneously exceeds 65\% on both answerable and unanswerable accuracy, revealing a miscalibration of current multimodal reasoners.}}}
    \label{fig:mm-aqa_overview}
    \vspace{-8.5mm}
\end{figure}

\noindent\textbf{Formalizing Unanswerability and EA.} We adopt a working definition scoped to this evaluation setting. Let $x = (I, q)$ denote a multimodal input-query pair with answer space $\mathcal{Y}$. An instance is \emph{answerable} if at least one $y^* \in \mathcal{Y}$ is unambiguously supported by the available evidence, and \emph{unanswerable} if no such answer can be reliably determined, including cases where multiple competing answers are equally defensible, since committing to any single response would be epistemically unjustified. A model \emph{abstains} by outputting a designated token (e.g., ``I don't know''), which is correct iff the instance is unanswerable.

\noindent\textbf{Limitations of existing work.}
Multimodal benchmarks like MMMU~\citep{yue2024mmmumassivemultidisciplinemultimodal} assume answerability, and while MMLongBench-Doc~\citep{ma2024mmlongbenchdocbenchmarkinglongcontextdocument}, and concurrent works like ~\citep{guo2024unkvqa,zhu2026mohobench,akter2024visreas} include unanswerable samples, it relies on coarse methods that do not reflect the full range of realistic failure modes. Robustness suites such as MM-Vet~\citep{yu2024mmvet} presuppose a valid answer always exists, and the abstention survey by \citet{wen-etal-2025-know} does not transfer to multimodal settings. Section~\ref{sec:related_work} provides a full comparison.

\noindent\textbf{Our Approach.} We introduce \textbf{MM-AQA} 
(\textbf{M}ulti\textbf{M}odal \textbf{A}bstention 
\textbf{Q}uestion \textbf{A}nswering), a benchmark and evaluation framework for studying abstention in multimodal reasoning, with the following \textit{key contributions}: \textbf{(1)} \textbf{MM-AQA}, an abstention-aware benchmark built from MMMU and MMLongBench-Doc, distinguished from concurrent works by a principled transformation taxonomy, dual-axis construction, and joint evaluation of VLMs and MAS;
\textbf{(2)} Structured, question-aware transformations along modality dependency and evidence sufficiency axes that convert answerable instances into unanswerable ones, requiring reasoning beyond surface cues;
\textbf{(3)} An automated construction pipeline with dual-consensus VLM verification and human validation to ensure dataset reliability;
\textbf{(4)} An evaluation framework spanning standalone VLMs and a three agent MAS built on AutoGen~\citep{wu2023autogen}, integrating verify-then-abstain~\citep{hu2019readverify} and multi-agent debate~\citep{du2024multiagentdebate}; coverage of six baselines from degenerate anchors to self-consistency and P(True);
\textbf{(5)} Empirical evidence of poor calibration and limited abstention in frontier models, revealing a clear accuracy–abstention Pareto frontier; \textbf{(6)}~Public release of datasets and infrastructure to support reproducible research on multimodal abstention (\emph{upon acceptance)}. Figure \ref{fig:mm-aqa_overview} gives an overview.

\section{Related Work}
\label{sec:related_work}
\noindent\textbf{Unanswerable question paradigms.}
SQuAD 2.0~\citep{rajpurkar2018squad2} operationalized evidence insufficiency 
detection by pairing answerable with adversarially unanswerable questions; we extend this to multimodal evidence. VizWiz~\citep{gurari2018vizwiz} introduced natural unanswerability via blind photographers. \citet{hu2019readverify} established the verify-then-abstain paradigm in 
reading comprehension, paralleling our multi-agent architecture.

\noindent\textbf{Selective prediction and abstention.}
\citet{el-yaniv2010selective} formalized the risk-coverage trade-off; 
\citet{geifman2017selective} extended it to deep networks. \citet{guo2017calibration} showed models are poorly calibrated, motivating MM-AQA's calibration metrics. 
\citet{madhusudhan-etal-2025-llms} introduced Abstain-QA and the AUCM confusion 
matrix, a methodological precursor to our protocol. 
\citet{kirichenko2025abstentionbench} find that reasoning fine-tuning degrades 
abstention and model scale has almost no effect, findings our experiments 
corroborate.

\noindent\textbf{LLM self-knowledge and uncertainty.}
\citet{yin2023largelanguagemodelsknow} established that models lag humans at identifying unanswerable queries; \citet{kadavath2022language} showed LLMs
can self-assess via P(True); \citet{tian2023calibration} that verbal confidence is better calibrated than raw probabilities; and \citet{xiong2024llmsexpressuncertainty} that models remain systematically overconfident. \citet{wang2023selfconsistency} and \citet{kuhn2023semantic} explore majority-vote
consistency and semantic entropy methods, respectively, though \citet{kapoor2024llmstaught} show both require fine-tuning to transfer reliably.

\noindent\textbf{Reliable VQA, multimodal abstention, and hallucination.}
\citet{whitehead2022reliable} established VQA abstention as selective 
prediction; \citet{guo2024unkvqa} introduce UNK-VQA with five perturbation 
types on VQA v2; \citet{zhu2026mohobench} propose MoHoBench evaluating 28 
MLLMs on 12K+ unanswerable questions, the most directly comparable concurrent 
benchmark. \citet{groot2024overconfidence} find vision models are overconfident 
with high calibration error. On hallucination, CHAIR~\citep{rohrbach2018hallucination} and 
POPE~\citep{pope2023} established standard hallucination metrics; 
HallusionBench~\citep{guan2024hallusionbench} shows frontier models remain 
highly prone to it; and \citet{farquhar2024semantic} show uncertainty estimation 
can detect hallucinations at scale, connecting the two failure modes we target.

\noindent\textbf{Multi-agent reasoning systems.}
Multi-agent debate~\citep{du2024multiagentdebate} showed agent disagreement 
improves factuality; \citet{zheng2023judging} validates LLM-as-judge at 
$>$80\% human agreement, the core justification for our Verifier design. 
\citet{feng2024donthallucinateabstain} improves abstention by 19.3\% via knowledge-gap probing. \citet{brahman2024artofsayingno} taxonomize 
noncompliance into incomplete, unsupported, and indeterminate requests; MM-AQA 
extends this to multimodal settings with instances from real benchmarks and multi-agent evaluation.

\section{Problem Formulation}
We consider a multimodal QA setting where each instance is $x = (I, q)$, with evidence $I$ and question $q$. We distinguish between \emph{answerable} instances $\mathcal{D}_A$, where $I$ contains sufficient, unambiguous information to determine $y^*$, and \emph{unanswerable} instances $\mathcal{D}_U$, where evidence is perturbed such that the required information is missing, ambiguous, contradictory, or out of scope, thus precluding any reliable determination of $y^*$. Note that $\mathcal{D}_U$ encompasses: (i) instances where no valid answer exists, and (ii) instances where multiple competing answers are equally defensible (e.g., adversarial ambiguity). Both cases warrant abstention, since committing to any single answer would be epistemically unjustified.
\newline\noindent Given input $x$, a model $f_\theta$ must either produce an answer or abstain:
\[
f_\theta(x) \in \mathcal{Y} \cup \{\text{ABSTAIN}\},
\quad\text{with}\quad
\begin{cases}
f_\theta(x) = y^* & \text{if } x \in \mathcal{D}_A, \\
f_\theta(x) = \text{ABSTAIN} & \text{if } x \in \mathcal{D}_U.
\end{cases}
\]
This induces a joint objective: maximizing answer accuracy on $\mathcal{D}_A$ while correctly
abstaining on $\mathcal{D}_U$, and introduces an inherent trade off between predictive performance and uncertainty awareness~\citep{geifman2017selective,whitehead2022reliable}.

\section{MM-AQA Benchmark}
\textbf{MM-AQA} (MultiModal - Abstention Question Answering) comprises 2,079 samples for systematically evaluating abstention in multimodal reasoning systems. It has two subsets: \textbf{Abstain-MMLongBench-Doc} (A-MMLBD, 1526 samples), and \textbf{Abstain-MMMU}
(A-MMMU, 553 samples), derived from MMLongBench-Doc~\citep{ma2024mmlongbenchdocbenchmarkinglongcontextdocument} and STEM domains of MMMU~\citep{yue2024mmmumassivemultidisciplinemultimodal}, respectively. The size disparity reflects the source benchmarks in that MMMU's validation split yields fewer questions. Both subsets maintain an approximate 1:1 answerable-to-unanswerable ratio.
 
The design principle is that unanswerability should arise from \emph{structural perturbations} targeting the evidence required to answer each question, and not superficial image degradation or incidental omission. Each transformed instance satisfies: (i) preservation of original task structure and domain complexity (ii) realistic failure modes (missing evidence, corruption, contradiction, inferential impossibility) and (iii) resistance to recovery via external knowledge or surface level pattern matching. These properties are enforced by a pipeline combining question-aware transformation routing, LLM-guided dependency analysis, and a Dual-Consensus-VLM Quality Control (DC-QC) module. After quality filtering, 12\% from A-MMMU and 17\% from A-MMLBD instances were rejected; more details in Appendix~\ref{app:dc-qc}. 

\noindent\textbf{Transformation Taxonomy and Selection.} We introduce a principled transformation taxonomy, for the construction of unanswerable instances, along two complementary axes: (i) \textbf{modality dependency} for MMMU-style tasks, where information is distributed across relatively separable visual and textual channels, and (ii) \textbf{evidence sufficiency} for MMLongBench-Doc-style settings, where answers must be derived from heterogeneous, distributed document evidence. To avoid trivial artifacts, we adopt \textbf{question-aware routing}: each instance is analyzed via LLM or heuristic classifier and routed to transformations that directly disrupt its specific reasoning pathway. Transformation distributions for unanswerable samples are shown in Figure~\ref{fig:combined_tran_distribution}; further details about taxonomies and selection are in Appendix~\ref{app:transformation_sel_bal_p-qc}; 

\begin{figure}[t]
    \centering
    \begin{subfigure}[t]{0.48\columnwidth}
        \centering
        \includegraphics[width=\linewidth]{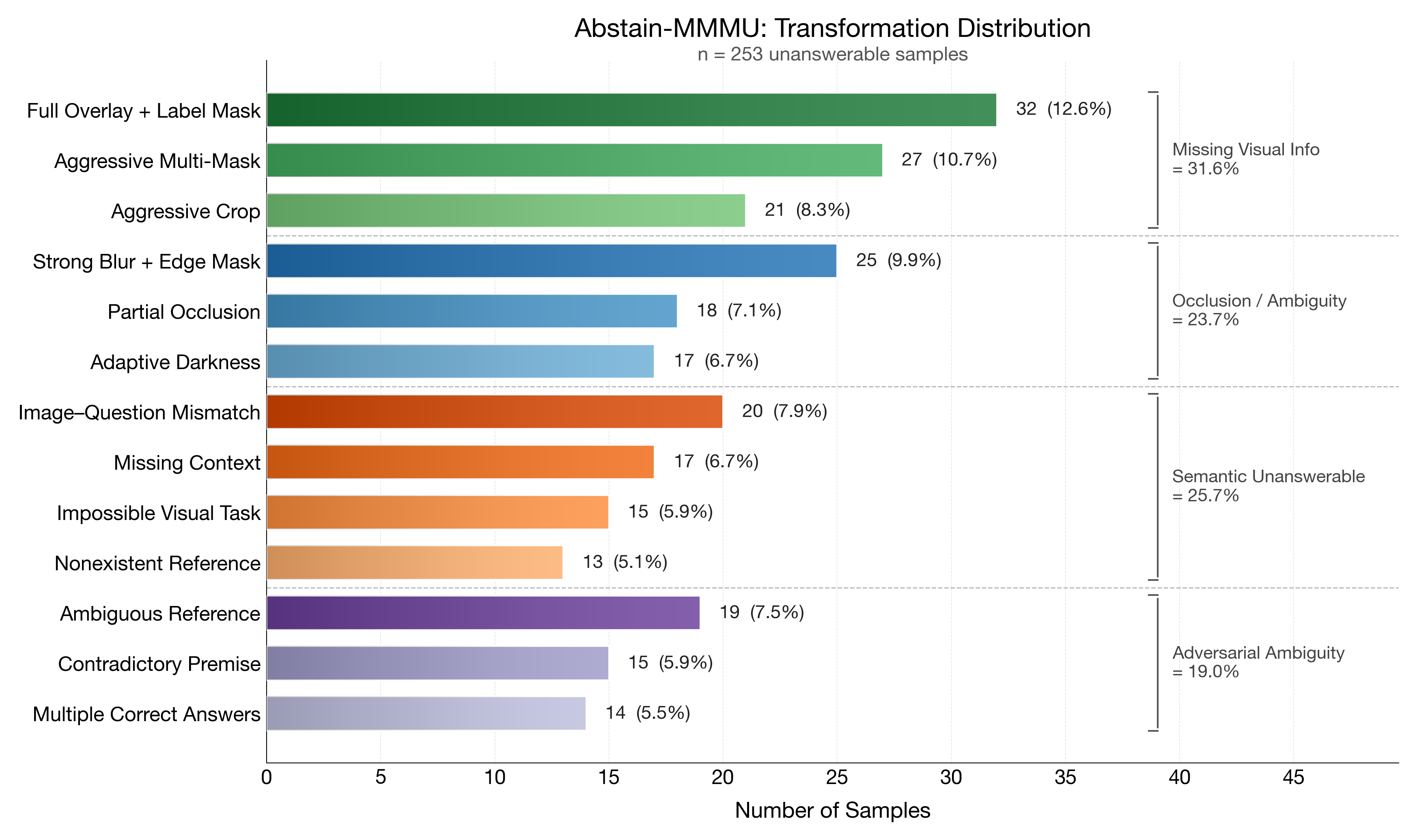}
        \label{fig:tran_distr_taxon_abs_mmmu}
    \end{subfigure}
    \hfill
    \begin{subfigure}[t]{0.48\columnwidth}
        \centering
        \includegraphics[width=\linewidth]{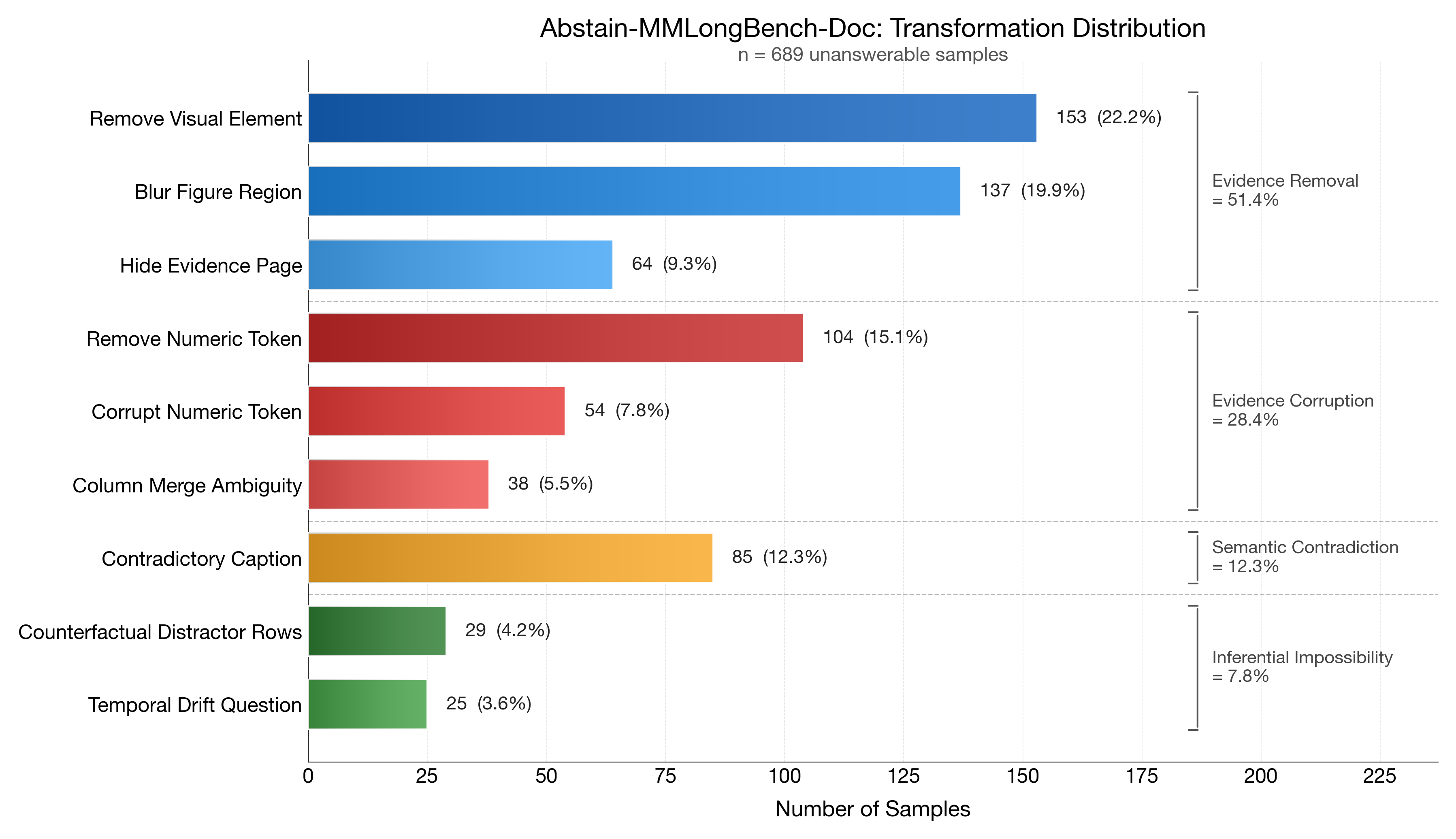}
        \label{fig:tran_distr_taxon_abs_mmlbd}
    \end{subfigure}
    \vspace{-3mm}
    \caption{\footnotesize {Taxonomy and transformation distribution for unanswerable samples across both benchmarks. \textbf{Left:} Abstain-MMMU; \textbf{Right:} Abstain-MMLongBench-Doc}}
    \label{fig:combined_tran_distribution}
    \vspace{-4mm}
\end{figure}

\noindent\textbf{Abstain-MMMU} is constructed from 10 STEM subsets of MMMU (Math, Physics, Biology, Chemistry, Geography, Mechanical Engineering, Computer Science, Clinical Medicine, Electronics and Materials), selected for their questions where visual evidence is the primary bottleneck for answerability. A-MMMU adopts a modality dependency paradigm where instances are classified into four image dependency levels (Decorative, Partial, Essential, Exclusive) enabling reliable routing: image transformations target Essential/Exclusive cases, text-level perturbations target Decorative/Partial. Transformations fall into four families: (1)~\textit{Missing Visual Information} (structured masking, aggressive cropping); (2)~\textit{Occlusion Ambiguity} (blur, extreme underexposure, occlusion patterns); (3)~\textit{Semantic Unanswerability} (question rewriting to require unavailable information); and (4)~\textit{Adversarial Ambiguity} (introducing multiple valid interpretations via contradictions or constraint removal).

\begin{figure}[t]
    \centering
    \includegraphics[width=1\textwidth]{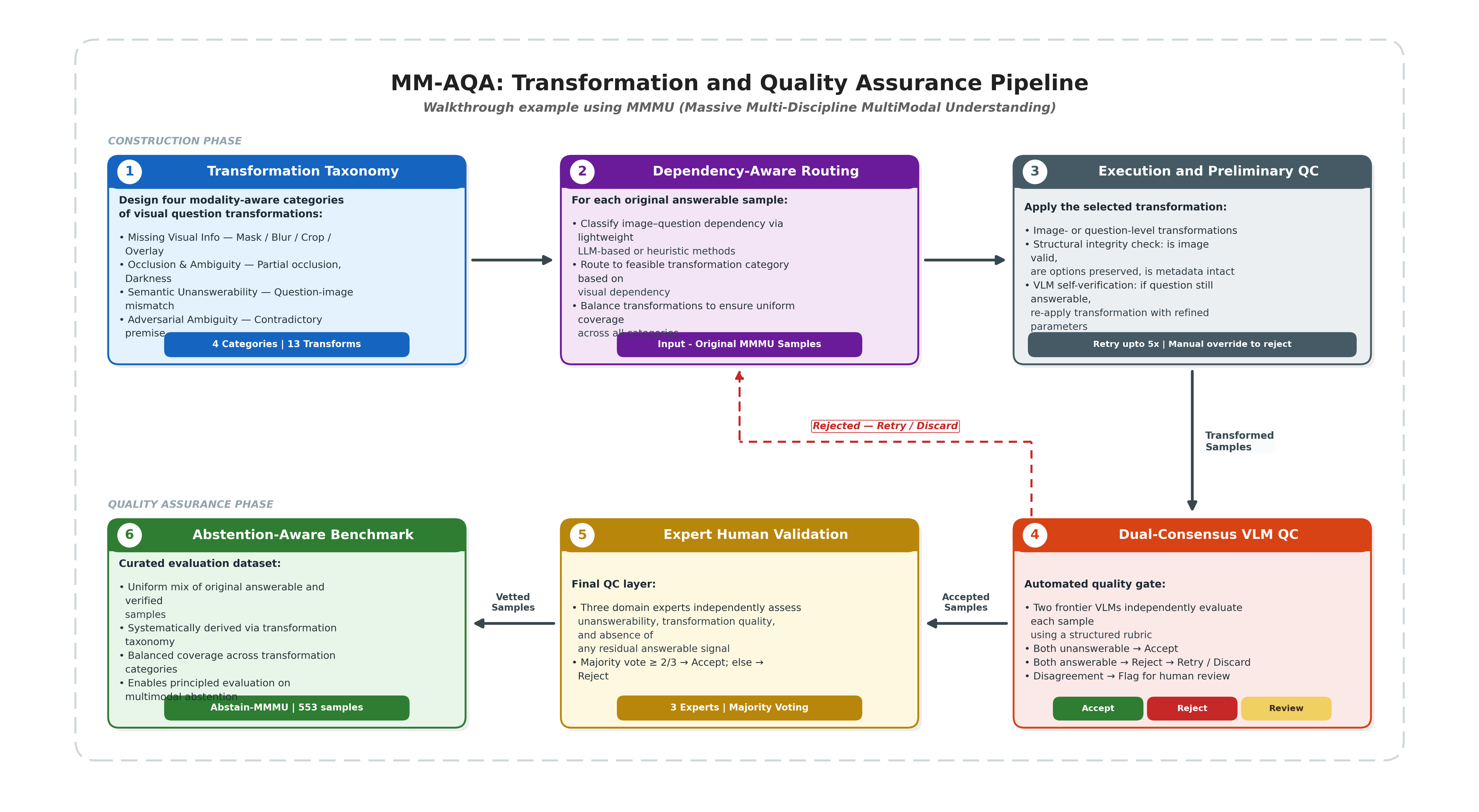}
     \vspace{-5mm}
    \caption{\footnotesize {\textbf{A walkthrough example of the MM-AQA pipeline}, illustrated via Abstain-MMMU curation; Abstain-MMLongBench-Doc follows the same process.}}
    \label{fig:tran_pipeline_mmmu}
    \vspace{-5mm}
\end{figure}

\noindent\textbf{Abstain-MMLongBench-Doc}, derived from MMLongBench-Doc, adopts an evidence-sufficiency paradigm targeting multi-page document reasoning. Questions are classified into seven types (Extractive, Counting, Aggregation, Comparative, Visual, Descriptive, Temporal) for evidence-aware routing. Transformation categories are: (1)~\textit{Evidence Removal} (OCR-guided masking, page-level hiding) (2)~\textit{Evidence Corruption} (numeric perturbation, table column distortion) (3)~\textit{Semantic Contradiction} (LLM-generated contradictory captions, unit/currency swaps) and (4)~\textit{Inferential Impossibility} (temporal shifts outside document coverage). A ranked fallback chain ensures robustness under rendering or OCR failures. Full details on transformation categories, all 22 transformations, selection, balancing, and preliminary quality verification for both datasets are in Appendix~\ref{app:transformation_catalogue}, \ref{app:transformation_sel_bal_p-qc}.

\noindent\textbf{Dual-Consensus-VLM Quality Control Module. }
Two independent VLMs perform QC by jointly
evaluating each transformed instance. A sample is \textit{rejected} if both models answer with justification (retried up to five times), \textit{accepted} if both agree on unanswerability, and \textit{flagged} for manual review on disagreement. This module uses VLMs from the same model family as those later evaluated; this is an acknowledged limitation. \textbf{Expert Human validation}, provides a quality signal not subject to the same circularity: Three annotators via majority voting, assess robustness of the DC-QC module and overall data quality, by evaluating a stratified sample of 200 instances per subset, drawn proportionally across transformation categories; Annotation guidelines and protocol are in Appendix~\ref{app:dc-qc}.

\section{Evaluation Methodology}
\label{sec:evaluation_methodology}

\noindent\textbf{Inference architectures.}
In the \textit{standalone} setting, a single VLM processes the full multimodal prompt. In the \textit{multi-agent} setting, we employ a three agent architecture: the \textbf{Reasoner} generates a candidate answer
with justification; the \textbf{Verifier} evaluates and outputs one of \emph{Approve} (accept the answer), \emph{Abstain} (override with 'I Don't Know'), or \emph{Request Revision} (feedback to improve the answer; only in non terminal rounds of iterative mode); the \textbf{Orchestrator} manages agent communication. We instantiate MAS in two modes: \textit{sequential}- single Reasoner-Verifier pass and \textit{iterative}- up to three Reasoner-Verifier rounds; abstention issued if disagreement persists at the final round.
More details in Appendix \ref{app:mas_config}.
 
\noindent\textbf{Experimental conditions and abstention clauses.} We consider three orthogonal experiment conditions: \textbf{(1)} \texttt{base}: models output only an answer with brief explanation, \textbf{(2)} \texttt{verbal confidence (vconf)}: models additionally provide a verbal confidence score on a discrete 1-5 scale, enabling threshold-based abstention~\citep{xiong2024llmsexpressuncertainty};
we prefer a coarser scale over the 0-100 of the original work
to make the threshold sweep ($\tau \in \{1,2,3,4\}$) exhaustive; and \textbf{(3)} \texttt{Chain of Thought (CoT)}: models are required to generate explicit step-by-step reasoning before the final answer. We also incorporate two
Abstention Clause variants ~\citep{madhusudhan-etal-2025-llms}:
\texttt{standard}: models are free to abstain but are not probed explicitly, and \texttt{extreme abstain}: applies graded
behavioral pressure with explicit consequences for incorrect responses. This yields a $3 \times 2$ design (condition $\times$ clause). More details on our experimental setup, condition and abstention templates, full prompt structure are detailed in Appendix~\ref{app:prompt_templates}; representative dataset samples from both A-MMMU and A-MMLBD are in Appendix~\ref{app:dataset_samples}.

\begin{figure}[t]
    \centering
    \includegraphics[width=1\textwidth]{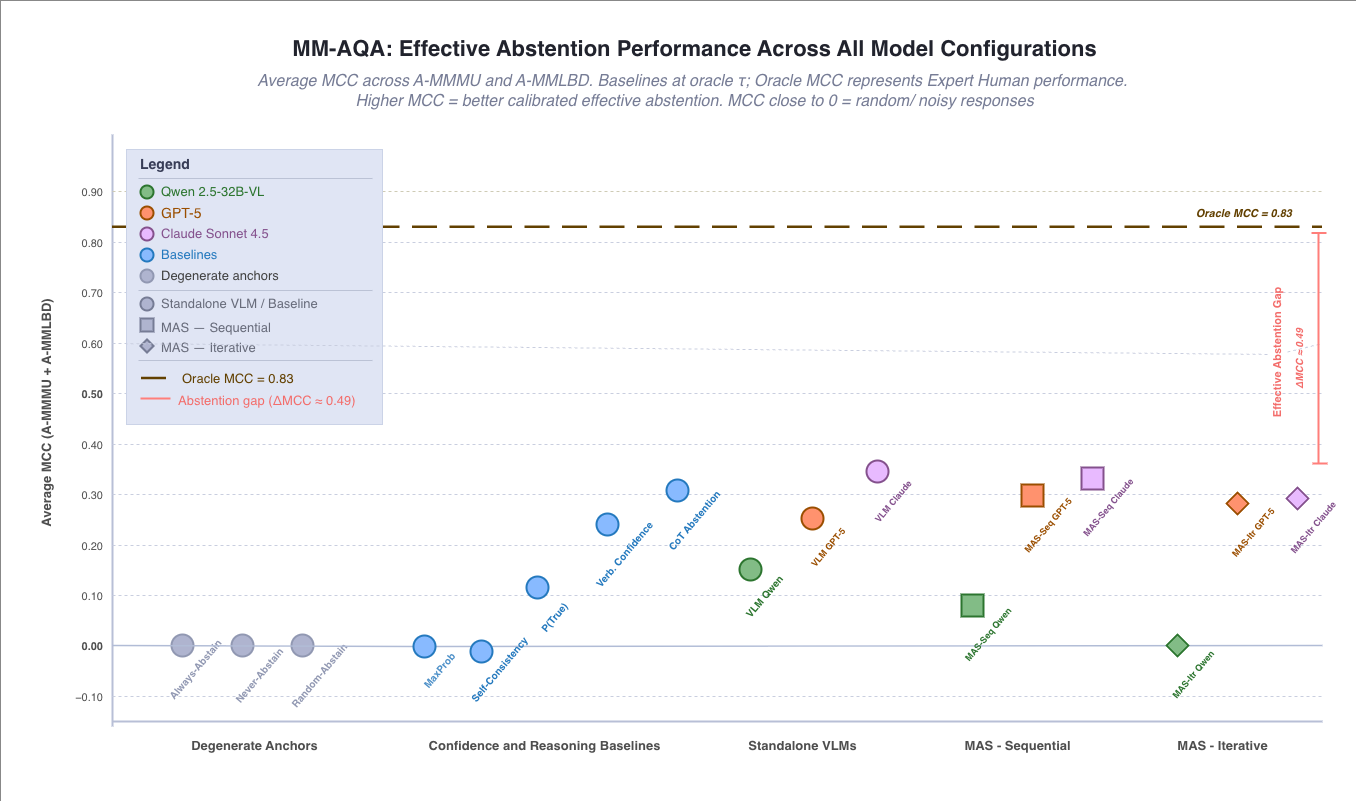}
\caption{
\footnotesize{\textbf{Abstention performance across all evaluated configurations on MM-AQA}
(avg MCC across A-MMMU and A-MMLBD; baselines at oracle~$\tau$).
Markers denote system type: \textbf{circles} = standalone VLM/ baseline,
\textbf{squares} = MAS-Sequential, \textbf{diamonds} = MAS-Iterative.
Colors denote model: \textbf{green} = Qwen~2.5-32B-VL, \textbf{orange} = GPT-5,
\textbf{purple} = Claude Sonnet~4.5, \textbf{blue} = confidence and reasoning
baselines, \textbf{gray} = degenerate anchors.
The brown dashed line marks the estimated human expert oracle
(MCC~$\approx$~0.83; Appendix~\ref{app:oracle_mcc_derivation});
the red bracket shows the gap ($\Delta\text{MCC} \approx 0.49$) to the best evaluated system (VLM Claude Sonnet~4.5, avg MCC~$=~0.344$). This confirms that MM-AQA poses a substantial and unresolved challenge for current frontier VLMs and MAS architectures.
}}
    \label{fig:mcc_comparison}
    \vspace{-5mm}
\end{figure}

\noindent\textbf{Abstention baselines.}
To contextualize multi-agent and prompting results against established methods, we include: \textit{Always/Never/Random-Abstain}, degenerate anchors; \textit{MaxProb Thresholding}~\citep{geifman2017selective}, on open-source VLMs; \textit{Verbal Confidence Thresholding}~\citep{xiong2024llmsexpressuncertainty}, via \texttt{vconf} predictions, (no additional inference);
\textit{Self-Consistency Abstention}~\citep{wang2023selfconsistency}, sample $N{=}10$ for each question, then abstain when no answer achieves majority; \textit{P(True) Self Evaluation}~\citep{kadavath2022language}, abstain when model's self-assessed correctness $< \tau$; and
\textit{CoT-Based Abstention Prompting}, CoT reasoning about evidence availability, before committing. Further details in Appendix \ref{app:subsec:baselines}.
 
\noindent\textbf{Confusion matrix and performance metrics.}
Building on ~\citet{madhusudhan-etal-2025-llms}, each response is mapped to a five-way confusion matrix conditioned on ground-truth answerability. Abstention is determined by case-insensitive substring matching against a wide predefined set of abstention indicators. Correctness on answered instances uses original benchmarks' gold labels and dataset-specific heuristics. The confusion matrix categories are as follows- True Positive: correct answer on
answerable, False Positive: incorrect answer on answerable, False Negative: abstention on answerable, True Negative: abstention on unanswerable, and AU or Answer on Unanswerable, with $\text{TP}+\text{FP}+\text{FN}+\text{TN}+\text{AU}=N$;
AU isolates hallucination under insufficient evidence from regular FP errors. Four metrics are computed: AAC $= \frac{\text{TP}}{\text{TP}+\text{FP}+\text{FN}}$,
UAC $= \frac{\text{TN}}{\text{TN}+\text{AU}}$,
AR $= \frac{\text{FN}+\text{TN}}{N}$, and MCC extended to the
five-category matrix by treating AU as a second false-positive
class:
$\text{MCC} = \frac{\text{TP}{\cdot}\text{TN}-(\text{FP}+\text{AU}){\cdot}\text{FN}}{\sqrt{(\text{TP}+\text{FP}+\text{AU})(\text{TP}+\text{FN})(\text{TN}+\text{FP}+\text{AU})(\text{TN}+\text{FN})}}$
(reduces to standard binary MCC when $\text{AU}{=}0$).
An interpretive framework is in Appendix~\ref{app:subsec:metric_interpretation}.

\section{Results and Discussion}
\label{sec:res_and_disc}
 
We evaluate three frontier VLMs, GPT-5, Claude Sonnet~4.5, and Qwen~2.5-32B-VL,
across standalone and MAS configurations, yielding nine system-model combinations
per dataset. We refer to these as 'GPT-5', 'Sonnet 4.5', 'Qwen 2.5' respectively. All main experiments are zero-shot, at $T{=}0.1$; Self-Consistency uses
$T{=}0.7$, $N{=}10$; P(True) and MaxProb use $T{=}0.1$ (Appendix \ref{app:evaluation_framework}).
For the \texttt{vconf} condition, we apply verbal confidence thresholding: a response
is treated as an abstention when the model's self-reported confidence score is $\leq \tau$. We swept $\tau \in \{1, 2, 3, 4\}$ and found $\tau{=}3$ yields
the best combined MCC across both datasets.
Tables~\ref{tab:main_mmmu} and~\ref{tab:main_mmlbd} report AAC, UAC, AR, and Best MCC
for every system-model-condition-clause combination on A-MMMU and A-MMLBD,
respectively. Table~\ref{tab:baselines} reports all abstention baselines.
Figure~\ref{fig:mcc_comparison} plots average MCC across both subsets against
the estimated human expert oracle (MCC~$\approx$~0.83), revealing that no configuration closes even half the gap to expert level abstention. 
\textit{\textbf{Oracle reference:}} The human expert oracle is \emph{derived}, not measured: human STEM accuracy on A-MMMU is extrapolated from frontier-model per-discipline ratios and published MMMU human
scores~\citep{yue2024mmmumassivemultidisciplinemultimodal};
UAC is set conservatively to 98.5\%. The $\Delta\text{MCC}\approx 0.49$ gap is robust across the plausible oracle range (0.46-0.52), confirming our qualitative conclusion; full derivation in Appendix~\ref{app:oracle_mcc_derivation}.
Our experiments yield the following key findings:

\begin{table}[t]
\centering
\small
\setlength{\tabcolsep}{4pt}
\renewcommand{\arraystretch}{1.05}
\begin{adjustbox}{width=\textwidth}
\begin{tabular}{llc | ccc | ccc | c}
\toprule
 
\multirow{2}{*}{\textbf{System}} &
\multirow{2}{*}{\textbf{Model}} &
\multirow{2}{*}{\textbf{Cond.}} &
\multicolumn{3}{c|}{\textbf{Standard clause}} &
\multicolumn{3}{c|}{\textbf{Extreme Abstain clause}} &
\multirow{2}{*}{\textbf{Best MCC}} \\
 
\cmidrule(lr){4-6}\cmidrule(lr){7-9}
& & &
AAC & UAC & AR &
AAC & UAC & AR &
\\
 
\midrule
\multicolumn{10}{l}{\textit{Single VLM}} \\
\midrule
 
\multirow{3}{*}{VLM}
  & \multirow{3}{*}{Claude Sonnet~4.5}
  & Base  & 60.7\se{2.8} & 1.2\se{0.7} & 0.5\se{0.3} & 59.7\se{2.8} & 41.7\se{3.1} & 21.6\se{1.8} & \multirow{3}{*}{0.262} \\
& & vconf & 57.7\se{2.9} & \textbf{26.6}\se{2.8} & 16.3\se{1.6} & 56.7\se{2.9} & \textbf{56.7}\se{3.1} & 33.9\se{2.0} &  \\
& & CoT   & \textbf{68.0}\se{2.7} & 1.6\se{0.8} & 0.7\se{0.4} & \textbf{60.7}\se{2.8} & 47.6\se{3.1} & 25.7\se{1.9} &  \\

\multirow{3}{*}{VLM}
  & \multirow{3}{*}{GPT-5}
  & Base  & \textbf{68.3}\se{2.7} & 4.9\se{1.4} & 3.0\se{0.7} & 61.8\se{2.8} & 37.2\se{3.0} & 23.0\se{1.8} & \multirow{3}{*}{0.197} \\
& & vconf & 62.8\se{2.8} & \textbf{39.0}\se{3.1} & 25.2\se{1.9} & 60.8\se{2.8} & \textbf{42.3}\se{3.1} & 26.4\se{1.9} &  \\
& & CoT   & 64.1\se{2.8} & 4.8\se{1.4} & 2.6\se{0.7} & \textbf{63.0}\se{2.8} & 39.8\se{3.1} & 23.3\se{1.8} &  \\

\multirow{3}{*}{VLM}
  & \multirow{3}{*}{Qwen~2.5-32B-VL}
  & Base  & \textbf{50.0}\se{2.9} & 4.0\se{1.2} & 2.2\se{0.6} & 47.3\se{2.9} & 18.2\se{2.4} & 9.4\se{1.2} & \multirow{3}{*}{0.111} \\
& & vconf & 49.0\se{2.9} & \textbf{8.3}\se{1.7} & 4.9\se{0.9} & 46.7\se{2.9} & \textbf{22.5}\se{2.6} & 12.1\se{1.4} &  \\
& & CoT   & 46.7\se{2.9} & 2.8\se{1.0} & 1.4\se{0.5} & \textbf{49.3}\se{2.9} & 17.4\se{2.4} & 9.8\se{1.3} &  \\

\midrule
\multicolumn{10}{l}{\textit{MAS - Sequential}} \\
\midrule
 
\multirow{3}{*}{MAS-Seq}
  & \multirow{3}{*}{Claude Sonnet~4.5}
  & Base  & \textbf{38.2}\se{2.8} & 82.9\se{2.4} & 65.0\se{2.0} & 36.3\se{2.8} & 90.1\se{1.9} & 68.5\se{2.0} & \multirow{3}{*}{\textbf{0.287}} \\
& & vconf & 38.0\se{2.8} & 84.5\se{2.3} & 65.8\se{2.0} & 36.4\se{2.8} & 90.1\se{1.9} & 68.5\se{2.0} &  \\
& & CoT   & 37.0\se{2.8} & \textbf{88.1}\se{2.0} & 68.3\se{2.0} & \textbf{39.0}\se{2.8} & \textbf{92.1}\se{1.7} & 70.5\se{1.9} &  \\

\multirow{3}{*}{MAS-Seq}
  & \multirow{3}{*}{GPT-5}
  & Base  & 53.6\se{2.9} & 61.7\se{3.1} & 40.1\se{2.1} & 57.1\se{2.9} & 64.0\se{3.0} & 41.9\se{2.1} & \multirow{3}{*}{0.278} \\
& & vconf & 53.7\se{2.9} & \textbf{65.2}\se{3.0} & 42.7\se{2.1} & 56.4\se{2.9} & \textbf{66.0}\se{3.0} & 43.4\se{2.1} &  \\
& & CoT   & \textbf{59.1}\se{2.9} & 59.3\se{3.1} & 37.3\se{2.1} & \textbf{59.1}\se{2.9} & 64.4\se{3.0} & 39.3\se{2.1} &  \\

\multirow{3}{*}{MAS-Seq}
  & \multirow{3}{*}{Qwen~2.5-32B-VL}
  & Base  & \textbf{36.0}\se{2.8} & 56.2\se{3.1} & 41.2\se{2.1} & 35.7\se{2.8} & 59.3\se{3.1} & 43.0\se{2.1} & \multirow{3}{*}{-0.005} \\
& & vconf & \textbf{36.0}\se{2.8} & \textbf{56.5}\se{3.1} & 41.6\se{2.1} & 35.7\se{2.8} & \textbf{60.1}\se{3.1} & 43.4\se{2.1} &  \\
& & CoT   & 35.7\se{2.8} & 53.0\se{3.1} & 38.2\se{2.1} & \textbf{38.3}\se{2.8} & 58.1\se{3.1} & 41.6\se{2.1} &  \\

\midrule
\multicolumn{10}{l}{\textit{MAS - Iterative (up to 3 rounds)}} \\
\midrule
 
\multirow{3}{*}{MAS-Itr}
  & \multirow{3}{*}{Claude Sonnet~4.5}
  & Base  & \textbf{39.3}\se{2.8} & 77.8\se{2.6} & 62.0\se{2.1} & 37.7\se{2.8} & 88.9\se{2.0} & 67.9\se{2.0} & \multirow{3}{*}{0.224} \\
& & vconf & 38.7\se{2.8} & \textbf{80.2}\se{2.5} & 63.6\se{2.0} & 37.4\se{2.8} & \textbf{89.3}\se{1.9} & 68.1\se{2.0} &  \\
& & CoT   & 38.7\se{2.8} & 77.4\se{2.6} & 63.2\se{2.1} & \textbf{40.3}\se{2.8} & 84.5\se{2.3} & 65.2\se{2.0} &  \\

\multirow{3}{*}{MAS-Itr}
  & \multirow{3}{*}{GPT-5}
  & Base  & 59.1\se{2.9} & 45.8\se{3.1} & 27.5\se{1.9} & \textbf{61.8}\se{2.8} & 53.0\se{3.1} & 30.4\se{2.0} & \multirow{3}{*}{0.265} \\
& & vconf & 56.1\se{2.9} & \textbf{52.6}\se{3.1} & 33.5\se{2.0} & 60.8\se{2.8} & \textbf{57.3}\se{3.1} & 34.1\se{2.0} &  \\
& & CoT   & \textbf{60.8}\se{2.8} & 42.3\se{3.1} & 24.8\se{1.8} & 61.1\se{2.8} & 48.2\se{3.1} & 29.5\se{1.9} &  \\

\multirow{3}{*}{MAS-Itr}
  & \multirow{3}{*}{Qwen~2.5-32B-VL}
  & Base  & \textbf{33.7}\se{2.7} & 70.0\se{2.9} & 52.1\se{2.1} & 33.2\se{2.7} & 65.2\se{3.0} & 49.9\se{2.1} & \multirow{3}{*}{-0.002} \\
& & vconf & \textbf{33.7}\se{2.7} & \textbf{70.4}\se{2.9} & 52.3\se{2.1} & \textbf{33.3}\se{2.7} & 65.2\se{3.0} & 50.1\se{2.1} &  \\
& & CoT   & 32.7\se{2.7} & \textbf{70.4}\se{2.9} & 51.0\se{2.1} & 32.3\se{2.7} & \textbf{69.6}\se{2.9} & 49.2\se{2.1} &  \\

\bottomrule
\end{tabular}
\end{adjustbox}
\caption{\footnotesize{
Results on \textbf{Abstain-MMMU}. AAC, UAC, AR, and Best MCC (highest MCC across all abstention clause-condition variants for each system-model pair) are reported. Standard errors are computed as $\sqrt{p(1-p)/n}$ where $n$ is the subset size. $n$ $\rightarrow$ AAC=300; UAC=253; AR=553.
}}
\label{tab:main_mmmu}
\vspace{-5mm}
\end{table}

\noindent\textbf{\textit{VLMs rarely abstain under standard prompting.}}
All three VLMs show near-zero abstention on unanswerable
instances across both datasets.
On A-MMMU, Sonnet~4.5 achieves 1.2\% UAC
(AR 0.5\%), GPT-5 achieves 4.9\% UAC, and Qwen~2.5 achieves 4.0\% UAC,
with MCC scores of 0.036-0.052 confirming near-random discrimination.
A-MMLBD shows modestly higher base UAC (Sonnet 4.5: 17.0\%, GPT-5: 2.0\%,
Qwen 2.5: 3.6\%), attributable to document-level formatting cues that occasionally surface evidence gaps. Yet all three models still answer incorrectly on more than 80\% of unanswerable instances, confirming that frontier VLMs tend to hallucinate across modalities and evidence types.

\begin{table}[t]
\centering
\small
\setlength{\tabcolsep}{4pt}
\renewcommand{\arraystretch}{1.05}
\begin{adjustbox}{width=\textwidth}
\begin{tabular}{llc | ccc | ccc | c}
\toprule
 
\multirow{2}{*}{\textbf{System}} &
\multirow{2}{*}{\textbf{Model}} &
\multirow{2}{*}{\textbf{Cond.}} &
\multicolumn{3}{c|}{\textbf{Standard clause}} &
\multicolumn{3}{c|}{\textbf{Extreme Abstain clause}} &
\multirow{2}{*}{\textbf{Best MCC}} \\
 
\cmidrule(lr){4-6}\cmidrule(lr){7-9}
& & &
AAC & UAC & AR &
AAC & UAC & AR &
\\
 
\midrule
\multicolumn{10}{l}{\textit{Single VLM}} \\
\midrule
 
\multirow{3}{*}{VLM}
  & \multirow{3}{*}{Claude Sonnet~4.5}
  & Base  & 67.3\se{1.6} & 17.0\se{1.4} & 8.6\se{0.7} & \textbf{64.6}\se{1.7} & 71.2\se{1.7} & 37.3\se{1.2} & \multirow{3}{*}{\textbf{0.430}} \\
& & vconf & 66.2\se{1.6} & \textbf{52.8}\se{1.9} & 27.1\se{1.1} & 63.8\se{1.7} & \textbf{75.7}\se{1.6} & 40.6\se{1.3} &  \\
& & CoT   & \textbf{67.8}\se{1.6} & 15.7\se{1.4} & 8.1\se{0.7} & 64.1\se{1.7} & 74.7\se{1.7} & 39.7\se{1.3} &  \\

\multirow{3}{*}{VLM}
  & \multirow{3}{*}{GPT-5}
  & Base  & \textbf{62.9}\se{1.7} & 2.0\se{0.5} & 1.2\se{0.3} & \textbf{61.2}\se{1.7} & 53.5\se{1.9} & 27.9\se{1.2} & \multirow{3}{*}{0.331} \\
& & vconf & 61.7\se{1.7} & \textbf{58.6}\se{1.9} & 31.2\se{1.2} & 61.1\se{1.7} & \textbf{59.0}\se{1.9} & 31.2\se{1.2} &  \\
& & CoT   & 62.4\se{1.7} & 0.6\se{0.3} & 0.6\se{0.2} & 59.9\se{1.7} & 51.4\se{1.9} & 26.8\se{1.1} &  \\

\multirow{3}{*}{VLM}
  & \multirow{3}{*}{Qwen~2.5-32B-VL}
  & Base  & 58.2\se{1.7} & 3.6\se{0.7} & 2.2\se{0.4} & \textbf{59.8}\se{1.7} & 37.9\se{1.8} & 20.9\se{1.1} & \multirow{3}{*}{0.217} \\
& & vconf & 58.0\se{1.7} & \textbf{13.6}\se{1.3} & 7.3\se{0.7} & 59.6\se{1.7} & \textbf{39.8}\se{1.9} & \textbf{22.0}\se{1.1} &  \\
& & CoT   & \textbf{59.6}\se{1.7} & 4.8\se{0.8} & 2.8\se{0.4} & 58.7\se{1.7} & 35.8\se{1.8} & 19.5\se{1.0} &  \\

\midrule
\multicolumn{10}{l}{\textit{MAS - Sequential}} \\
\midrule
 
\multirow{3}{*}{MAS-Seq}
  & \multirow{3}{*}{Claude Sonnet~4.5}
  & Base  & \textbf{46.2}\se{1.7} & 85.2\se{1.4} & 61.7\se{1.2} & 45.4\se{1.7} & 94.3\se{0.9} & 67.3\se{1.2} & \multirow{3}{*}{0.383} \\
& & vconf & \textbf{46.2}\se{1.7} & \textbf{90.7}\se{1.1} & 64.3\se{1.2} & 45.3\se{1.7} & 94.3\se{0.9} & 67.4\se{1.2} &  \\
& & CoT   & 45.6\se{1.7} & 88.1\se{1.2} & 62.8\se{1.2} & \textbf{46.2}\se{1.7} & \textbf{95.2}\se{0.8} & 66.8\se{1.2} &  \\

\multirow{3}{*}{MAS-Seq}
  & \multirow{3}{*}{GPT-5}
  & Base  & 55.8\se{1.7} & 69.4\se{1.8} & 42.7\se{1.3} & \textbf{57.3}\se{1.7} & 74.5\se{1.7} & 45.0\se{1.3} & \multirow{3}{*}{0.321} \\
& & vconf & 55.6\se{1.7} & \textbf{76.3}\se{1.6} & 46.6\se{1.3} & 56.8\se{1.7} & \textbf{75.8}\se{1.6} & 46.3\se{1.3} &  \\
& & CoT   & \textbf{56.3}\se{1.7} & 68.1\se{1.8} & 40.8\se{1.3} & 57.1\se{1.7} & 73.3\se{1.7} & 43.2\se{1.3} &  \\

\multirow{3}{*}{MAS-Seq}
  & \multirow{3}{*}{Qwen~2.5-32B-VL}
  & Base  & 45.0\se{1.7} & 64.4\se{1.8} & 46.4\se{1.3} & 42.7\se{1.7} & 74.6\se{1.7} & 53.4\se{1.3} & \multirow{3}{*}{0.171} \\
& & vconf & 45.1\se{1.7} & 65.9\se{1.8} & 47.1\se{1.3} & 42.9\se{1.7} & 74.6\se{1.7} & 53.4\se{1.3} &  \\
& & CoT   & \textbf{46.8}\se{1.7} & \textbf{67.5}\se{1.8} & 47.2\se{1.3} & \textbf{43.1}\se{1.7} & \textbf{78.5}\se{1.6} & 54.2\se{1.3} &  \\

\midrule
\multicolumn{10}{l}{\textit{MAS - Iterative (up to 3 rounds)}} \\
\midrule
 
\multirow{3}{*}{MAS-Itr}
  & \multirow{3}{*}{Claude Sonnet~4.5}
  & Base  & \textbf{46.1}\se{1.7} & 83.0\se{1.4} & 60.6\se{1.3} & 45.4\se{1.7} & \textbf{95.1}\se{0.8} & 66.3\se{1.2} & \multirow{3}{*}{0.367} \\
& & vconf & 46.0\se{1.7} & \textbf{91.1}\se{1.1} & 64.4\se{1.2} & 45.3\se{1.7} & \textbf{95.1}\se{0.8} & 66.3\se{1.2} &  \\
& & CoT   & 46.0\se{1.7} & 84.1\se{1.4} & 59.7\se{1.3} & \textbf{46.7}\se{1.7} & 93.3\se{1.0} & 65.6\se{1.2} &  \\

\multirow{3}{*}{MAS-Itr}
  & \multirow{3}{*}{GPT-5}
  & Base  & 60.1\se{1.7} & 48.9\se{1.9} & 27.4\se{1.1} & 59.4\se{1.7} & 62.7\se{1.8} & 35.3\se{1.2} & \multirow{3}{*}{0.303} \\
& & vconf & 58.9\se{1.7} & \textbf{63.8}\se{1.8} & 36.3\se{1.2} & 58.7\se{1.7} & \textbf{66.0}\se{1.8} & 37.8\se{1.2} &  \\
& & CoT   & \textbf{60.8}\se{1.7} & 46.8\se{1.9} & 26.0\se{1.1} & \textbf{60.2}\se{1.7} & 58.8\se{1.9} & 32.8\se{1.2} &  \\

\multirow{3}{*}{MAS-Itr}
  & \multirow{3}{*}{Qwen~2.5-32B-VL}
  & Base  & \textbf{50.1}\se{2.5} & 58.6\se{1.9} & 45.4\se{1.5} & 45.3\se{1.7} & 72.4\se{1.7} & 49.5\se{1.3} & \multirow{3}{*}{0.200} \\
& & vconf & 50.0\se{2.5} & 59.7\se{1.9} & 46.3\se{1.5} & 45.4\se{1.7} & 72.6\se{1.7} & 49.5\se{1.3} &  \\
& & CoT   & 49.2\se{1.7} & \textbf{63.4}\se{1.8} & 42.6\se{1.3} & \textbf{47.1}\se{1.7} & \textbf{75.9}\se{1.6} & 51.2\se{1.3} &  \\

\bottomrule
\end{tabular}
\end{adjustbox}
\caption{
\footnotesize Results on \textbf{Abstain-MMLongBench-Doc}. AAC, UAC, AR, and Best MCC (highest MCC across all abstention clause-condition variants for each system-model pair) are reported.
Standard errors are computed as $\sqrt{p(1-p)/n}$ where $n$ is the subset size. $n$ $\rightarrow$ AAC=837; UAC=689; AR=1526.
}
\label{tab:main_mmlbd}
\vspace{-5mm}
\end{table}

\noindent\textbf{\textit{Baselines reveal a confidence-signal gap.}}
Table~\ref{tab:baselines} reports all abstention baselines (VLM Sonnet~4.5); verbalized confidence thresholding yields MCC of 0.113 on A-MMMU ($\tau{=}1$) and 0.368 on A-MMLBD ($\tau{=}4$), outperforming standard prompting.
The divergent thresholds expose a critical asymmetry: on A-MMMU the model assigns
uniformly high confidence regardless of answerability, confirming verbalized confidence
is a weak signal on perceptual tasks.
CoT Abstention achieves the strongest single-model results: 64.3\% AAC / 62.8\% UAC
(MCC 0.334) on A-MMMU, and 46.4\% AAC / 82.8\% UAC (MCC 0.284) on A-MMLBD.
Self-Consistency yields negative MCC on A-MMMU ($-$0.109) and modest MCC on A-MMLBD
(0.083), confirming response variance is not a reliable signal for perceptual tasks.
P(True) achieves modest MCC (${\approx}0.11$-0.12) on both benchmarks; MaxProb
degenerates to abstain-always on both (Appendix~\ref{app:subsec:baselines}).
On A-MMLBD, CoT Abstention (MCC 0.284) falls below MAS (0.380-0.383); 
on A-MMMU it exceeds MAS (0.334 vs.\ 0.287), indicating MAS overhead is most justified for document reasoning tasks.

\noindent\textbf{\textit{Abstention clauses improve UAC but do not resolve miscalibration.}}
Introducing the extreme abstain clause substantially lifts UAC across all models
on both datasets.
On A-MMMU: Sonnet~4.5 improves from 1.2\% to 41.5\%; GPT-5 from 4.9\% to 37.2\%.
On A-MMLBD: Sonnet~4.5 improves from 17.0\% to 71.2\%; GPT-5 from 2.0\% to 53.5\%;
Qwen 2.5 from 3.6\% to 37.9\%. AAC degrades by only 1-5\% across both datasets.
Yet no standalone model exceeds 75.7\% UAC under any prompting variant, and all
still answer incorrectly on a majority of unanswerable queries.
The large UAC jump under extreme abstain clause reveals that latent uncertainty
signals exist but are suppressed under default decoding, consistent with
text-only abstention findings~\citep{madhusudhan-etal-2025-llms}, and suggesting
fine-tuning on graded examples is needed rather than prompting
alone~\citep{kapoor2024llmstaught}.

\noindent\textbf{\textit{MAS improves abstention but introduces an 
accuracy-abstention trade-off.}} The Verifier drives abstention in MAS, 
overriding the Reasoner in 27-52\% of cases. On A-MMMU, Claude-based MAS 
achieves 75-92\% UAC at the cost of reduced AAC (36-40\%), while GPT-5-based 
MAS strikes the most balanced trade-off: MAS-Seq GPT-5 (CoT, exabs) achieves 
58.7\% AAC / 64.4\% UAC (MCC 0.274). On A-MMLBD, the pattern sharpens: 
Sonnet~4.5 MAS-Seq achieves up to 95.2\% UAC but suppresses AAC to 45-46\%; 
GPT-5 MAS-Seq yields 57.3\% AAC / 74.5\% UAC (MCC 0.321). Across both 
datasets, no system simultaneously exceeds 65\% on both AAC and UAC, and 
removing the Verifier reverts UAC to single-digit performance, confirming 
that abstention in MAS is enforced rather than emergent.
 
\begin{table}[t]
\centering
\small
\setlength{\tabcolsep}{4pt}
\renewcommand{\arraystretch}{1.05}
\begin{adjustbox}{width=\textwidth}
\begin{tabular}{lc | ccc | ccc}
\toprule
 
\multirow{2}{*}{\textbf{Baseline}} &
\multirow{2}{*}{\textbf{Variant}} &
\multicolumn{3}{c|}{\textbf{Abstain-MMMU}} &
\multicolumn{3}{c}{\textbf{Abstain-MMLongBench}} \\
 
\cmidrule(lr){3-5}\cmidrule(lr){6-8}
& & AAC & UAC & MCC & AAC & UAC & MCC \\
 
\midrule
\multicolumn{8}{l}{\textit{Degenerate anchors}} \\
\midrule
 
Always-Abstain  & --- &  0.0  & 100.0 & \multicolumn{1}{c|}{---} &  0.0  & 100.0 & --- \\
Never-Abstain   & --- & 60.6  &   0.0 & \multicolumn{1}{c|}{---} & 66.7  &   0.0 & --- \\
Random-Abstain  & --- & 30.0  &  49.0 & \multicolumn{1}{c|}{0.000} & 35.6 &  50.2 & 0.000 \\
 
\midrule
\multicolumn{8}{l}{\textit{Confidence and reasoning-based}} \\
\midrule
 
Verb.\ Confidence$^{\ddagger}$
    & --- & 60.7\se{2.8} &  5.5\se{1.4} & \multicolumn{1}{c|}{0.113} & \textbf{60.0}\se{1.7} & 76.6\se{1.6} & \textbf{0.368} \\
 
CoT Abstention
    & --- & \textbf{64.3}\se{2.8} & 62.8\se{3.0} & \multicolumn{1}{c|}{\textbf{0.334}} & 46.4\se{1.7} & 82.8\se{1.4} & 0.284 \\
 
Self-Consistency
    & --- & 49.0\se{2.9} & 17.3\se{2.4} & \multicolumn{1}{c|}{$-$0.109} & 41.4\se{1.7} & 69.2\se{1.8} & 0.083 \\
 
P(True) self-eval.\
    & --- & 29.3\se{0.0} & 84.9\se{0.0} & \multicolumn{1}{c|}{0.111} & 35.1\se{0.0} & 84.7\se{0.0} & 0.119 \\
 
MaxProb$^{\dagger}$
    & Qwen~2.5-32B-VL & 0.0\se{0.0} & \textbf{100.0}\se{0.0} & \multicolumn{1}{c|}{0.000} & 0.0\se{0.0} & \textbf{100.0}\se{0.0} & 0.000 \\
 
\bottomrule
\end{tabular}
\end{adjustbox}
\caption{
  \footnotesize{\textbf{Abstention baselines (Sonnet~4.5 unless noted).}
  MCC at oracle~$\tau$.
  \textit{Verb.\ Confidence} thresholds the 1-5 score from the \texttt{vconf}
  condition; oracle $\tau{=}1$ (A-MMMU) / $\tau{=}4$ (A-MMLBD).
  \textit{CoT Abstention}: single forward pass, $T{=}0.1$, standard clause.
  \textit{Self-Consistency}: $N{=}10$, $T{=}0.7$; abstains when no answer achieves majority.
  \textit{P(True)}: self-assessed correctness probability; oracle $\tau{=}0.855$
  (A-MMMU) / $\tau{=}0.905$ (A-MMLBD); abstains below threshold.
  \textit{MaxProb} (oracle $\tau{=}0.98$ on A-MMMU, $\tau{=}0.99$ on A-MMLBD)
  degenerates to abstain-always at oracle~$\tau$
  (Appendix~\ref{app:subsec:baselines}).
  $^{\dagger}$Token-level logits required; unavailable for proprietary APIs.
  $^{\ddagger}$Oracle $\tau$ is an upper bound on the same test set.
  Standard errors: $\sqrt{p(1-p)/n}$; $n{=}300/837$ (AAC), $n{=}253/689$ (UAC),
  for A-MMMU/A-MMLBD.}}
\label{tab:baselines}
\vspace{-6mm}
\end{table}

\noindent\textbf{\textit{More iterative rounds do not help.}} On A-MMMU, 
MAS-Seq Sonnet~4.5 achieves MCC 0.283 vs.\ 0.221 for MAS-Itr. On A-MMLBD, 
both are comparable (0.380 vs.\ 0.363 for Sonnet~4.5; 0.321 vs.\ 0.303 for 
GPT-5). GPT-5 MAS-Itr recovers AAC on A-MMLBD (60.1\% vs.\ 55.6\%) but 
loses UAC. Qwen~2.5 MAS configurations underperform their standalone baseline on both subsets (MAS-Seq/MAS-Itr Best MCC $-$0.005/$-$0.002 vs.\ 0.111 on A-MMMU; 0.171/0.200 vs.\ 0.217 on A-MMLBD): when the Reasoner is weakly 
calibrated, additional rounds amplify rather than correct miscalibration.

\noindent\textbf{\textit{Models tracks structural explicitness, not modality.}}
Per-transformation results reveal a consistent within-dataset pattern.
On A-MMMU, \textit{Semantic Unanswerability} yields the highest UAC (78.5\%, $+$31pp
above mean) while \textit{Missing Visual Information} yields the lowest (26.2\%,
$-$21pp), despite both involving an unchanged modality paired with a disrupted one.
On A-MMLBD, \textit{Evidence Removal} is easiest to detect (86.4\%, $+$12pp) while
\textit{Semantic Contradiction} is hardest (37.6\%, $-$37pp).
Absolute UAC values should not be compared across benchmarks given the $\sim$27pp gap
in baseline abstention tendencies (A-MMMU: 47.4\%; A-MMLBD: 74.5\%).
The unifying principle is \textit{structural explicitness}: models abstain when
unanswerability is clearly encoded, a broken question, an absent page; but
attempt recovery when evidence is physically present yet degraded. More details in Appendix~\ref{app:per_transform_analysis}.

\section{Conclusion}
\label{sec:conclusion}
We introduce MM-AQA, a multimodal abstention benchmark of 2079 samples, constructed via 22 transformation types along two complementary axes. Across three frontier
VLMs and two multi-agent architectures, a
consistent picture emerges; under standard prompting, frontier models hallucinate answers on nearly all unanswerable instances. Multi-agent verification substantially improves abstention but introduces a clear
accuracy-abstention trade-off that no configuration resolves. Critically, the bottleneck is not
reasoning depth as sequential multi-agent designs match or exceed iterative variants. The sensitivity of UAC
to extreme abstention clause (a jump from 1\% to 41\%) reveals that models suppress latent uncertainty signals rather than lacking them. These findings motivate three directions: (1) abstention-aware
training is necessary to move beyond the
identified Pareto frontier; (2) uncertainty estimation methods designed for the multimodal
setting like semantic entropy over visual inputs, VLM specific calibration, remain largely unexplored; (3) MM-AQA's pipeline should be extended to non-STEM domains, video, and multi-document settings.

\bibliography{colm2026_conference}

@misc{yue2024mmmumassivemultidisciplinemultimodal,
      title={MMMU: A Massive Multi-discipline Multimodal Understanding and Reasoning Benchmark for Expert AGI}, 
      author={Xiang Yue and Yuansheng Ni and Kai Zhang and Tianyu Zheng and Ruoqi Liu and Ge Zhang and Samuel Stevens and Dongfu Jiang and Weiming Ren and Yuxuan Sun and Cong Wei and Botao Yu and Ruibin Yuan and Renliang Sun and Ming Yin and Boyuan Zheng and Zhenzhu Yang and Yibo Liu and Wenhao Huang and Huan Sun and Yu Su and Wenhu Chen},
      year={2024},
      eprint={2311.16502},
      archivePrefix={arXiv},
      primaryClass={cs.CL},
      url={https://arxiv.org/abs/2311.16502}, 
}

@misc{ma2024mmlongbenchdocbenchmarkinglongcontextdocument,
      title={MMLongBench-Doc: Benchmarking Long-context Document Understanding with Visualizations}, 
      author={Yubo Ma and Yuhang Zang and Liangyu Chen and Meiqi Chen and Yizhu Jiao and Xinze Li and Xinyuan Lu and Ziyu Liu and Yan Ma and Xiaoyi Dong and Pan Zhang and Liangming Pan and Yu-Gang Jiang and Jiaqi Wang and Yixin Cao and Aixin Sun},
      year={2024},
      eprint={2407.01523},
      archivePrefix={arXiv},
      primaryClass={cs.CV},
      url={https://arxiv.org/abs/2407.01523}, 
}

@article{el-yaniv2010selective,
  title={On the Foundations of Noise-Free Selective Classification},
  author={El-Yaniv, Ran},
  journal={Journal of Machine Learning Research},
  volume={11},
  pages={1605--1641},
  year={2010}
}

@article{kadavath2022language,
  title={Language Models (Mostly) Know What They Know},
  author={Kadavath, Saurav and others},
  journal={arXiv preprint arXiv:2207.05221},
  year={2022}
}

@article{yu2024mmvet,
  title={MM-Vet: Evaluating Large Multimodal Models for Integrated Capabilities},
  author={Yu, Wenliang and others},
  journal={arXiv preprint arXiv:2308.02490},
  year={2024}
}

@article{wu2023autogen,
  title={AutoGen: Enabling Next-Gen LLM Applications via Multi-Agent Conversation Framework},
  author={Wu, Qingyun and others},
  journal={arXiv preprint arXiv:2308.08155},
  year={2023}
}

@inproceedings{madhusudhan-etal-2025-llms,
    title = "Do {LLM}s Know When to {NOT} Answer? Investigating Abstention Abilities of Large Language Models",
    author = "Madhusudhan, Nishanth  and
      Madhusudhan, Sathwik Tejaswi  and
      Yadav, Vikas  and
      Hashemi, Masoud",
    editor = "Rambow, Owen  and
      Wanner, Leo  and
      Apidianaki, Marianna  and
      Al-Khalifa, Hend  and
      Eugenio, Barbara Di  and
      Schockaert, Steven",
    booktitle = "Proceedings of the 31st International Conference on Computational Linguistics",
    month = jan,
    year = "2025",
    address = "Abu Dhabi, UAE",
    publisher = "Association for Computational Linguistics",
    url = "https://aclanthology.org/2025.coling-main.627/",
    pages = "9329--9345"
}

@inproceedings{rajpurkar2018squad2,
   title={Know What You Don't Know: Unanswerable Questions for SQuAD},
   author={Rajpurkar, Pranav and Jia, Robin and Liang, Percy},
   booktitle={ACL},
   year={2018}
 }

@inproceedings{gurari2018vizwiz,
   title={{VizWiz} Grand Challenge: Answering Visual Questions from Blind People},
   author={Gurari, Danna and Li, Qing and Stangl, Abigale J and Guo, Anhong and Lin, Chi and Grauman, Kristen and Luo, Jiebo and Bigham, Jeffrey P},
   booktitle={CVPR},
   year={2018}
 }

@inproceedings{hu2019readverify,
   title={Read + Verify: Machine Reading Comprehension with Unanswerable Questions},
   author={Hu, Minghao and Wei, Furu and Peng, Yuxing and Huang, Zhen and Yang, Nan and Li, Dongsheng},
   booktitle={AAAI},
   year={2019}
 }

@inproceedings{whitehead2022reliable,
   title={Reliable Visual Question Answering: Abstain Rather Than Answer Incorrectly},
   author={Whitehead, Spencer and Wu, Suzanne and Farhadi, Ali and Krishna, Ranjay and Rastegari, Mohammad},
   booktitle={ECCV},
   year={2022}
 }

@inproceedings{geifman2017selective,
   title={Selective Classification for Deep Neural Networks},
   author={Geifman, Yonatan and El-Yaniv, Ran},
   booktitle={NeurIPS},
   year={2017}
 }

@inproceedings{guo2017calibration,
   title={On Calibration of Modern Neural Networks},
   author={Guo, Chuan and Pleiss, Geoff and Sun, Yu and Weinberger, Kilian Q},
   booktitle={ICML},
   year={2017}
 }

@inproceedings{yin2023largelanguagemodelsknow,
   title={Do Large Language Models Know What They Don't Know?},
   author={Yin, Zhangyue and Sun, Qiwei and Geng, Qipeng and Li, Jiannan and Dai, Xipeng and Qiu, Xipeng},
   booktitle={Findings of ACL},
   year={2023}
 }

@inproceedings{kuhn2023semantic,
   title={Semantic Uncertainty: Linguistic Invariances for Uncertainty Estimation in Natural Language Generation},
   author={Kuhn, Lorenz and Gal, Yarin and Farquhar, Sebastian},
   booktitle={ICLR},
   year={2023}
 }

@article{farquhar2024semantic,
   title={Detecting Hallucinations in Large Language Models Using Semantic Entropy},
   author={Farquhar, Sebastian and Kossen, Jannik and Kuhn, Lorenz and Gal, Yarin},
   journal={Nature},
   year={2024}
 }

@inproceedings{tian2023calibration,
   title={Just Ask for Calibration: Strategies for Eliciting Calibrated Confidence Scores from Language Models Fine-Tuned with Human Feedback},
   author={Tian, Katherine and Mitchell, Eric and Yao, Huaxiu and Manning, Christopher D and Finn, Chelsea},
   booktitle={EMNLP},
   year={2023}
 }

@misc{mathew2021docvqadatasetvqadocument,
      title={DocVQA: A Dataset for VQA on Document Images}, 
      author={Minesh Mathew and Dimosthenis Karatzas and C. V. Jawahar},
      year={2021},
      eprint={2007.00398},
      archivePrefix={arXiv},
      primaryClass={cs.CV},
      url={https://arxiv.org/abs/2007.00398}, 
}

@article{kwiatkowski-etal-2019-natural,
    title = "Natural Questions: A Benchmark for Question Answering Research",
    author = "Kwiatkowski, Tom  and
      Palomaki, Jennimaria  and
      Redfield, Olivia  and
      Collins, Michael  and
      Parikh, Ankur  and
      Alberti, Chris  and
      Epstein, Danielle  and
      Polosukhin, Illia  and
      Devlin, Jacob  and
      Lee, Kenton  and
      Toutanova, Kristina  and
      Jones, Llion  and
      Kelcey, Matthew  and
      Chang, Ming-Wei  and
      Dai, Andrew M.  and
      Uszkoreit, Jakob  and
      Le, Quoc  and
      Petrov, Slav",
    editor = "Lee, Lillian  and
      Johnson, Mark  and
      Roark, Brian  and
      Nenkova, Ani",
    journal = "Transactions of the Association for Computational Linguistics",
    volume = "7",
    year = "2019",
    address = "Cambridge, MA",
    publisher = "MIT Press",
    url = "https://aclanthology.org/Q19-1026/",
    doi = "10.1162/tacl_a_00276",
    pages = "452--466"
}

@inproceedings{kamath2020selective,
   title={Selective Question Answering under Domain Shift},
   author={Kamath, Amita and Jia, Robin and Liang, Percy},
   booktitle={ACL},
   year={2020}
 }

@inproceedings{xiong2024llmsexpressuncertainty,
   title={Can LLMs Express Their Uncertainty? An Empirical Evaluation of Confidence Elicitation in LLMs},
   author={Xiong, Miao and Hu, Zhiyuan and Lu, Xinyang and Li, Yifei and Fu, Jie and He, Junxian and Hooi, Bryan},
   booktitle={ICLR},
   year={2024}
 }

@inproceedings{groot2024overconfidence,
   title={Overconfidence is Key: Verbalized Uncertainty Evaluation in Large Language and Vision-Language Models},
   author={Groot, Tobias and Valdenegro-Toro, Matias},
   booktitle={TrustNLP@ACL},
   year={2024}
 }

@inproceedings{kapoor2024llmstaught,
   title={Large Language Models Must Be Taught to Know What They Don't Know},
   author={Kapoor, Sanyam and Gruver, Nate and Roberts, Manley and Collins, Katherine and Pal, Arka and Bhatt, Umang and Weller, Adrian and Wilson, Andrew Gordon and Steinhardt, Jacob and Neel, Seth},
   booktitle={NeurIPS},
   year={2024}
 }

@inproceedings{pope2023,
   title={Evaluating Object Hallucination in Large Vision-Language Models},
   author={Li, Yifan and Du, Yifan and Zhou, Kun and Wang, Jinpeng and Zhao, Wayne Xin and Wen, Ji-Rong},
   booktitle={EMNLP},
   year={2023}
 }

@inproceedings{rohrbach2018hallucination,
   title={Object Hallucination in Image Captioning},
   author={Rohrbach, Anna and Hendricks, Lisa Anne and Burns, Kaylee and Darrell, Trevor and Saenko, Kate},
   booktitle={EMNLP},
   year={2018}
 }

@inproceedings{guan2024hallusionbench,
   title={{HallusionBench}: An Advanced Diagnostic Suite for Entangled Language Hallucination and Visual Illusion in Large Vision-Language Models},
   author={Guan, Tianrui and Liu, Fuxiao and Wu, Xiyang and Xian, Ruiqi and Li, Zongxia and Liu, Xiaoyu and Wang, Xijun and Chen, Lijun and Huang, Fuwei and He, Heng and others},
   booktitle={CVPR},
   year={2024}
 }

@inproceedings{wang2023selfconsistency,
   title={Self-Consistency Improves Chain of Thought Reasoning in Language Models},
   author={Wang, Xuezhi and Wei, Jason and Schuurmans, Dale and Le, Quoc and Chi, Ed and Narang, Sharan and Chowdhery, Aakanksha and Zhou, Denny},
   booktitle={ICLR},
   year={2023}
 }

@inproceedings{du2024multiagentdebate,
   title={Improving Factuality and Reasoning in Language Models through Multiagent Debate},
   author={Du, Yilun and Li, Shuang and Torralba, Antonio and Tenenbaum, Joshua B and Mordatch, Igor},
   booktitle={ICML},
   year={2024}
 }

@inproceedings{zheng2023judging,
   title={Judging LLM-as-a-Judge with MT-Bench and Chatbot Arena},
   author={Zheng, Lianmin and Chiang, Wei-Lin and Sheng, Ying and Zhuang, Siyuan and Wu, Zhanghao and Zhuang, Yonghao and Lin, Zi and Li, Zhuohan and Li, Dacheng and Xing, Eric P and others},
   booktitle={NeurIPS},
   year={2023}
 }

@inproceedings{feng2024donthallucinateabstain,
   title={Don't Hallucinate, Abstain: Identifying LLM Knowledge Gaps via Multi-LLM Collaboration},
   author={Feng, Shangbin and Shi, Weijia and Wang, Yike and Ding, Wenxuan and Balachandran, Vidhisha and Tsvetkov, Yulia},
   booktitle={ACL},
   year={2024}
 }

@inproceedings{brahman2024artofsayingno,
   title={The Art of Saying No: Contextual Noncompliance in Language Models},
   author={Brahman, Faeze and Bisk, Yonatan and Choi, Yejin and Sap, Maarten},
   booktitle={NeurIPS},
   year={2024}
 }

@article{wen-etal-2025-know,
    title = "Know Your Limits: A Survey of Abstention in Large Language Models",
    author = "Wen, Bingbing  and
      Yao, Jihan  and
      Feng, Shangbin  and
      Xu, Chenjun  and
      Tsvetkov, Yulia  and
      Howe, Bill  and
      Wang, Lucy Lu",
    journal = "Transactions of the Association for Computational Linguistics",
    volume = "13",
    year = "2025",
    address = "Cambridge, MA",
    publisher = "MIT Press",
    url = "https://aclanthology.org/2025.tacl-1.26/",
    doi = "10.1162/tacl_a_00754",
    pages = "529--556"
}

@inproceedings{kirichenko2025abstentionbench,
   title={Reasoning LLMs Fail on Unanswerable Questions},
   author={Kirichenko, Polina and others},
   booktitle={NeurIPS},
   year={2025}
 }

@inproceedings{guo2024unkvqa,
   title={{UNK-VQA}: A Dataset and a Probe into the Abstention Ability of Multi-modal Large Language Models},
   author={Guo, Yan and Wang, Xinyu and Gao, Peng and Wang, Cairong},
   booktitle={ACM Multimedia},
   year={2024}
 }

@inproceedings{zhu2026mohobench,
   title={Assessing Honesty of Multimodal Large Language Models via Unanswerable Visual Questions},
   author={Zhu, Zhecan and others},
   booktitle={AAAI},
   year={2026}
 }

@inproceedings{akter2024visreas,
   title={{VISREAS}: Complex Visual Reasoning with Unanswerable Questions},
   author={Akter, Syeda Nahida and others},
   booktitle={Findings of ACL},
   year={2024}
 }

@inproceedings{srinivasan-etal-2024-selective,
    title = "Selective ``Selective Prediction'': Reducing Unnecessary Abstention in Vision-Language Reasoning",
    author = "Srinivasan, Tejas  and
      Hessel, Jack  and
      Gupta, Tanmay  and
      Lin, Bill Yuchen  and
      Choi, Yejin  and
      Thomason, Jesse  and
      Chandu, Khyathi",
    editor = "Ku, Lun-Wei  and
      Martins, Andre  and
      Srikumar, Vivek",
    booktitle = "Findings of the Association for Computational Linguistics: ACL 2024",
    month = aug,
    year = "2024",
    address = "Bangkok, Thailand",
    publisher = "Association for Computational Linguistics",
    url = "https://aclanthology.org/2024.findings-acl.767/",
    doi = "10.18653/v1/2024.findings-acl.767",
    pages = "12935--12948"
}
\bibliographystyle{colm2026_conference}
\appendix
\section{Appendix}

\section{Per-Transformation Analysis of MM-AQA}
\label{app:per_transform_analysis}

For this analysis we use the diagnostic configuration: standalone VLM, 
Claude Sonnet~4.5, ExAbs, CoT. Performance wise, it shows the most balance between AAC and UAC scores, enabling a neutral ground for per-transformation analysis. Therefore, we choose this model as it activates abstention without ceiling compression from over abstaining MAS setups.

\noindent\textbf{The structural explicitness principle.}
Per-category UAC is best predicted by how clearly unanswerability 
is encoded in the input structure, regardless of modality. 
A transformation has high structural explicitness if the model 
can identify evidence insufficiency without attempting answer 
recovery; low explicitness if the input appears complete or 
partially recoverable, drawing the model toward committing to 
an answer before the insufficiency becomes apparent.

\noindent\textbf{A-MMMU.}
\textit{Semantic Unanswerability} (78.5\%) is the easiest category 
because the question is self-evidently broken, referencing 
nonexistent entities or impossible visual tasks, while the image 
is unchanged. The model can identify insufficiency from the question 
text alone without any visual inference. \textit{Missing Visual 
Information} (26.2\%) is the hardest despite involving more severe 
evidence disruption: the question is unchanged and the image, 
though partially masked, is structurally present with surviving 
pixels in unmasked regions. The coherent question text exerts a 
strong answerable prior, drawing the model toward inference from 
image fragments rather than abstention. \textit{Occlusion Ambiguity} 
(43.3\%) sits above Missing Visual Information because extreme 
darkness or heavy uniform blur is itself a legible degradation 
signal, providing a weak cue that the image is irrecoverable. 
\textit{Adversarial Ambiguity} (45.8\%) is near the dataset mean; 
unlike Semantic Unanswerability, ambiguity does not make a question 
self-evidently unanswerable; the model can produce a locally 
coherent answer to one interpretation without recognizing that 
others are equally valid.

\noindent\textbf{A-MMLBD.}
\textit{Evidence Removal} (86.4\%) is the easiest category because 
hidden pages and 95\%-masked numeric tokens and 60\%-masked text tokens, produce a structurally salient gap. This allows the model to observe a document with missing content, thus identifying evidence insufficiency before attempting any computation. \textit{Evidence Corruption} (71.4\%) is harder: all pages and tables are present, but numeric values 
are corrupted and distractor rows inserted; the model partially 
attempts to compute answers from corrupted values before recognizing 
the insufficiency. \textit{Inferential Impossibility} (64.8\%) 
involves a temporal question rewrite analogous to A-MMMU's 
Semantic Unanswerability, yet UAC is lower because the structurally 
complete document invites the model to extrapolate trends rather 
than recognize the question as out of scope. \textit{Semantic 
Contradiction} (37.6\%, $-$37pp below mean) is the hardest 
category across both benchmarks relative to its dataset mean: 
a contradictory caption is planted alongside intact and correct 
document evidence and question, drawing the model toward reconciliation rather 
than abstention. This is the inverse of Evidence Removal in that, instead 
of an explicit absence, the model faces a conflict it 
attempts to resolve.

\begin{table}[t]
\centering
\footnotesize
\setlength{\tabcolsep}{7pt}
\renewcommand{\arraystretch}{1.05}
\begin{tabular}{llcc}
\toprule

\textbf{Dataset} & \textbf{Transformation category} & \textbf{N} & \textbf{UAC (\%)}\\

\midrule
\multicolumn{4}{l}{\textit{Abstain-MMMU}} \\
\midrule

& Missing Visual Information  & 80 & 26.2 \se{4.9}\\
& Occlusion Ambiguity         & 60 & 43.3 \se{6.4}\\
& Semantic Unanswerability    & 65 & 78.5 \se{5.1}\\
& Adversarial Ambiguity       & 48 & 45.8 \se{7.2}\\

\midrule
\multicolumn{4}{l}{\textit{Abstain-MMLongBench-Doc}} \\
\midrule

& Evidence Removal           & 354 & 86.4 \se{1.8}\\
& Evidence Corruption        & 196 & 71.4 \se{3.2}\\
& Semantic Contradiction     & 85 & 37.6 \se{5.3}\\
& Inferential Impossibility  & 54 & 64.8 \se{6.5}\\

\bottomrule
\end{tabular}
\caption{
\footnotesize \textbf{Per-transformation UAC} under a fixed diagnostic configuration
(standalone VLM, Claude Sonnet~4.5, ExAb, CoT), chosen to avoid 
ceiling compression from over-abstaining MAS setups. 
Within-dataset comparisons are appropriate; absolute values should 
not be compared across benchmarks given the $\sim$27pp gap in 
baseline abstention tendencies (A-MMMU: 47.4\%; A-MMLBD: 74.5\%). 
The consistent within-dataset pattern reflects \textit{structural 
explicitness}: categories where unanswerability is legibly encoded 
in input structure, broken question text, absent document 
pages, fall above their dataset mean UAC, while categories where 
evidence is present but degraded, corrupted, or contradictory fall 
below it. Standard errors: $\sqrt{p(1-p)/n}$; categories with 
N$<$100 carry higher uncertainty.
}
\label{tab:per_transform}
\end{table}

\noindent\textbf{Sample size caveats.}
Three categories carry elevated uncertainty: Adversarial Ambiguity 
(N=48, SE\,$\pm$7.2\%), Inferential Impossibility (N=54, 
SE\,$\pm$6.5\%), and Semantic Contradiction (N=85, SE\,$\pm$5.3\%). 
The broad pattern: Semantic Unanswerability substantially 
performs worse than image-degradation categories on A-MMMU; similarly,  Evidence Removal shows major degraded performance compared to Semantic Contradiction on A-MMLBD. This pattern is robust given the larger gaps and sample sizes involved (especially in A-MMLBD). Fine-grained orderings within the mid range categories are not to be over interpreted.
 
\section{MM-AQA Construction Details}
\label{app:mm-aqa_construction}
 
\subsection{Dataset Statistics and Source Distributions}
\label{app:subsec:dataset_statistics}
 
Table~\ref{tab:dataset_stats} summarizes dataset composition.
MMMU subsets were restricted to STEM domains (10 subjects in total) because they yield the highest proportion of Essential and Exclusive image-dependency questions. Extending coverage to Non-STEM subjects is left as future work.
The size disparity between benchmark subsets reflects source benchmark structure: we make use of the validation set of original MMMU benchmark which contains only 30 samples per subject, while MMLongBench-Doc did not contain a validation set, hence we used the entire train split.
Both subsets maintain an approximate 1:1 answerable-to-unanswerable ratio, ensuring performance differences reflect abstention behavior rather than class imbalance.
 
\begin{table}[t]
\centering
\small
\begin{tabular}{lrrr}
\toprule
\textbf{Subset} & \textbf{Total} & \textbf{Answerable} & \textbf{Unanswerable} \\
\midrule
A-MMMU     &  553 & 300 (54.2\%) & 253 (45.8\%) \\
A-MMLBD    & 1526 & 837 (54.8\%) & 689 (45.2\%) \\
\textbf{Total} & \textbf{2079} & \textbf{1137 (54.7\%)} & \textbf{942 (45.3\%)} \\
\bottomrule
\end{tabular}
\caption{MM-AQA dataset statistics by subset.}
\label{tab:dataset_stats}
\vspace{-6mm}
\end{table}
 
\subsection{Image Dependency Classification (A-MMMU)}
\label{app:subsec:dependency_classification}
 
Each question is classified into one of four dependency levels via a single LLM call (heuristics are used as fallback) evaluating
(i) whether all answer-relevant information is present in the question text, and (ii) whether
the image is strictly necessary for reasoning.
 
\noindent\textbf{Dependency levels.}
\textit{Decorative}: the image provides illustrative context only; the question is fully answerable from text alone.
\textit{Partial}: the image provides auxiliary cues that aid but are not required for answering.
\textit{Essential}: critical information exists only in the image; the question cannot be answered from text alone.
\textit{Exclusive}: the question is entirely image driven, with no supporting textual context.
 
\noindent\textbf{Heuristic fallback.}
When LLM access is unavailable/ fails, a rule-based classifier uses: (i) numeric token density (higher density increases text-sufficiency score), (ii) formula detection (mathematical expressions increase text-sufficiency likelihood), and (iii) visual keyword score (tokens such as ``graph'', ``diagram'', ``figure'' increase image-dependency score).
 
\noindent\textbf{Routing logic.}
Classification drives transformation routing: image transformations are applied only to Essential/Exclusive instances; text-level perturbations are applied to Decorative/Partial instances. This dependency-aware routing ensures transformations target the information source that is genuinely critical to answering, avoiding trivial artifacts. This routing is performed by a Transformation Selector Module which works entirely on rule base methods and heuristics.
 
\subsection{Evidence-Sufficiency Classification (A-MMLBD)}
\label{app:subsec:question-type_classification}
 Questions are classified into seven types using only rule-based signals. Detection patterns are listed in Table~\ref{tab:qtype_patterns}. We also employ a priority order, for situations where a question gets marked with multiple equally likely categories: Counting $\rightarrow$ Aggregation $\rightarrow$ Comparative $\rightarrow$ Visual $\rightarrow$ Extractive $\rightarrow$ Temporal $\rightarrow$ Descriptive.
 
This classification drives evidence-aware routing: Evidence Removal transformations are preferred for Extractive and Comparative questions (direct value lookup); Evidence Corruption targets Aggregation and Counting questions (numeric and text manipulation disrupts computation); Inferential Impossibility applies to Temporal questions; Semantic Contradiction is assigned primarily to Descriptive and Visual questions.
 
\begin{table}[ht]
\centering
\small
\begin{tabular}{ll}
\toprule
\textbf{Type} & \textbf{Trigger keywords / patterns} \\
\midrule
Counting      & ``how many'', ``number of'' \\
Aggregation   & ``sum'', ``total'', ``average'', ``mean'' \\
Comparative   & ``highest'', ``lowest'', ``difference between'', ``compared to'' \\
Visual        & ``chart'', ``graph'', ``figure'', ``image'', ``diagram'' \\
Temporal      & year/quarter patterns (e.g., ``FY2020'', ``Q3'', ``in \textit{year}'') \\
Extractive    & direct entity lookups (``what is'', ``who is'', ``name the'') \\
Descriptive   & description or summary requests (``describe'', ``explain'', ``summarize'') \\
\bottomrule
\end{tabular}
\caption{Evidence-sufficiency classification patterns for A-MMLBD routing.}
\label{tab:qtype_patterns}
\end{table}
 
\section{Transformation Catalogue}
\label{app:transformation_catalogue}
 
\subsection{A-MMMU Transformations}
\label{app:subsec:ammmu_transforms}
 
Thirteen unique transformations are organized into four families targeting complementary failure
modes. All image operations are applied programmatically via PIL/OpenCV; all text operations
are generated by a constrained LLM rewriter with downstream validation checks (output length,
semantic divergence, grammatical structure).
 
\noindent\textbf{(1) Missing Visual Information.}
Removes critical visual content via direct spatial manipulation.
\begin{enumerate}
  \item \textit{Aggressive crop}: removes the outermost 50\% from a randomly selected edge (top/bottom/left/right), eliminating axis labels, legends, or annotations depending on selection.
  \item \textit{Aggressive multi mask}: 20 randomly-placed black rectangles, each covering approximately 15\% of image area; enforces $\geq$70\% total coverage.
  \item \textit{Full overlay + Label mask}: semi-transparent gray overlay ($\alpha{\approx}0.6$) applied to the full image, followed by targeted edge masking of detected titles, captions, axis labels, and legends.
\end{enumerate}
 
\noindent\textbf{(2) Occlusion Ambiguity.}
Preserves image structure while blocking interpretability.
\begin{enumerate}\setcounter{enumi}{3}
  \item \textit{Partial occlusion}: 9 solid black bars (alternating horizontal and vertical, width $W/10$ to $W/5$) placed at regular intervals.
  \item \textit{Adaptive darkness}: all pixel values divided by 15, corresponding to approximately 3.9 stop exposure reduction; image remains non-empty but visually irrecoverable.
  \item \textit{Blur with edge mask}: Gaussian blur (radius 18) applied to $\approx$85\% of the image, followed by edge masking targeting axis labels and figure titles.
\end{enumerate}
 
\noindent\textbf{(3) Semantic Unanswerability.}
Modifies the question while leaving all images unchanged; requires non-trivial reasoning to
detect because image content is intact.
\begin{enumerate}\setcounter{enumi}{6}
  \item \textit{Image-question mismatch}: question is rewritten to ask about visual content that is entirely a different phenomenon the image cannot address.
  \item \textit{Missing context}: question is rewritten to require comparative, or extrinsic information unavailable from the image.
  \item \textit{Impossible visual task}: question is rewritten to require precise measurements, hidden properties, or sub-pixel distinctions not feasible from the image.
  \item \textit{Nonexistent reference}: a specific entity not present in the image is introduced into the question.
\end{enumerate}
All variants use constrained LLM rewriting with three validation gates: (i) output length within 120\% of input; (ii) semantic divergence confirmed via embedding distance; (iii) grammatical structure check via dependency parse.
 
\noindent\textbf{(4) Adversarial Ambiguity.}
Introduces multiple equally defensible interpretations, making commitment to any single answer epistemically unjustified.
\begin{enumerate}\setcounter{enumi}{10}
  \item \textit{Multiple correct answers}: question is reworded to introduce a conflicting constraint or alternative interpretive convention, such as specifying that a given parameter may be treated under two distinct standards, such that two or more answers options become simultaneously defensible.
  \item \textit{Ambiguous reference}: question is reinvented to ask about non-existent/ fabricated visual elements within the same domain.
  \item \textit{Contradictory premise}: question is reworded to introduce a premise that is logically inconsistent with one or more of the given conditions or constraints. The resulting contradiction makes it impossible to derive a self-consistent solution.
\end{enumerate}
 
\subsection{A-MMLBD Transformations}
\label{app:subsec:ammlbd_transforms}
 
Nine unique transformations organized into four families targeting evidence-structure disruption
in multi-page document settings.
 
\noindent\textbf{(1) Evidence Removal.}
Eliminates the specific document region required to answer the question.
\begin{enumerate}
  \item \textit{Remove visual element}: random black rectangular masks are placed in and around the center of the evidence pages where visual evidence density is highest.
  \item \textit{Blur figure region}: a heavy Gaussian blur is applied to the figure or chart region on the evidence page, rendering its visual features,  including axis labels, data trends, and color distinctions, unrecognizable. If OCR based feature detection fails, a fallback blur is applied to the entire page.
  \item \textit{Hide evidence page}: 50\% of the evidence pages of a question are hidden and the remaining 50\% are vertically cropped (at midpoint).
\end{enumerate}
 
\noindent\textbf{(2) Evidence Corruption.}
Preserves document structure while altering the semantic content required for answering.
\begin{enumerate}\setcounter{enumi}{3}
  \item \textit{Remove numeric token}: numeric values appearing in a table, chart, or text region of evidence pages are surgically erased (along with random text tokens). It leaves the surrounding context intact.
  \item \textit{Corrupt numeric token}: numeric values in a table, chart, or text region of evidence pages are surgically corrupted (along with erasure of random text tokens), introducing a subtle factual error that cannot be detected from the question text alone.
  \item \textit{Column merge ambiguity}: two or more table columns are merged into a single column with an ambiguous composite headers, obscuring the schema structure required to correctly interpret the tabular data.
\end{enumerate}
 
\noindent\textbf{(3) Semantic Contradiction.}
Injects conflicting information to create ambiguity.
\begin{enumerate}\setcounter{enumi}{6}
  \item \textit{Contradictory caption}: an LLM-generated note containing information that directly contradicts the question content is appended at the end.
\end{enumerate}
 
\noindent\textbf{(4) Inferential Impossibility.}
Modifies the question scope to require information outside the document's coverage.
\begin{enumerate}\setcounter{enumi}{7}
  \item \textit{Counterfactual Distractor Rows}: One or more plausible but factually incorrect rows are inserted into a table on the evidence pages, introducing distractor entries that share structural similarity with the ground-truth row.
  \item \textit{Temporal Drift}: the temporal reference in the question, such as a fiscal year or reporting period, is shifted to a time period outside the range covered by the document.
\end{enumerate}
 
\section{Transformation Selection, Balancing, and Preliminary Verification}
\label{app:transformation_sel_bal_p-qc}
 
\noindent\textbf{Scoring and routing.}
From both benchmark subsets, each transformation is scored against the question using a feature vector comprising: table/figure presence, quantitative reasoning required, cross-page dependency, visual keyword score, and
estimated computational complexity of the required reasoning step (this description is abridged; both subsets have different vectors with applicable features). Hard-constraint violations (e.g., image transformation assigned to a Decorative instance, or temporal drift applied to a non-temporal question) incur a $-95\%$ score penalty; soft violations (e.g., a transformation that is feasible but suboptimal for the question type) incur
$-30\%$. Transformations with score $>0.4$ are eligible; the threshold is relaxed to $>0.25$ if no candidate clears 0.4.
 
\noindent\textbf{Diversity balancing.}
After per-instance routing, a sequential balancing pass enforces near-uniform category distribution across the dataset while respecting per-instance feasibility constraints. The selector always prefers the least-used transformation category in the current pool, breaking ties by score. Note that the first priority is always appropriate transformation routing, next comes the balancing. This is the reason why we see slightly imbalanced transformation distributions (especially in Abstain-MMLBD).
 
\noindent\textbf{Fallback chain.}
For Abstain-MMLBD, a ranked fallback chain handles rendering or OCR failures at runtime. Terminal fallbacks, guaranteed to succeed on any valid image/ evidence page, are:
(a) full-page Gaussian blur (radius 25) and (b) content-guided dense masking ($\geq$80\% coverage).
 
\noindent\textbf{Preliminary verification.}
Before entering the Dual-Consensus QC module, each transformed instance, in both subsets, undergoes a preliminary
check: (i) structural integrity (image dimensions unchanged, no corrupt file headers, non-empty question text); (ii) a VLM self-verification loop that rejects instances answered with high confidence; (iii) a retry mechanism that explores up to three alternative transformations before discarding a failed instance. Acceptance requires ANSWERABLE with LOW confidence; MEDIUM or HIGH confidence triggers rejection.
 
\section{Quality Verification Details}
\label{app:dc-qc}
 
\noindent\textbf{Dual-Consensus VLM QC module.}
Post preliminary verification and complete dataset creation, two independent VLMs evaluate each transformed instance without access to the original
answerable version.
Decision rules:
\textit{Both UNANSWERABLE} $\rightarrow$ accepted;
\textit{Both ANSWERABLE with justification} $\rightarrow$ rejected (retry up to 5 times with refined transformation selection and parameters);
\textit{Disagreement} $\rightarrow$ flagged for mandatory human review.
We note that the QC module uses VLMs from the same model family as those later evaluated; human validation provides an independent quality signal not subject to this circularity.
 
\noindent\textbf{Rejection rates by category.}
Overall rejection rate: $\approx$15\% across both subsets.
In A-MMMU (12\% overall), the highest rejection rates occur in Adversarial Ambiguity (18\%) and Semantic Unanswerability (16\%), where residual answerability, where the question remains answerable despite the transformation, is most common. In A-MMLBD (17\% overall), Evidence Corruption has the highest rejection rate (22\%), as numeric perturbations of insufficient magnitude fail to block answer recovery via context; parameters were tightened ($\pm$20-50\% range) after observing this in preliminary runs. These rates inform future transformation calibration.
 
\noindent\textbf{Human annotation protocol.}
Three expert annotators (compensated at local wage standards) independently
evaluated a stratified random sample of 200 instances per subset (400 total), drawn proportionally across transformation categories.
The annotation task was designed to assess two properties simultaneously:
(i)~whether the DC-QC module incorrectly accepted instances that remained answerable (false
acceptances); and (ii)~overall data quality for both answerable and unanswerable-labeled instances.
 
To this end, annotators examined both categories of DC-QC output: all instances the module flagged as \textit{answerable} (and therefore rejected or requiring retry), and a stratified sample of instances the module flagged as \textit{unanswerable} (accepted into MM-AQA). For the latter, final judgments were obtained by majority vote; annotation guidelines defined
``unanswerable'' as requiring that no correct answer can be reliably determined from the provided evidence alone, without recourse to external knowledge.
 
\textbf{Outcomes.}
All unanswerable-labeled instances in the stratified sample were confirmed as genuinely unanswerable by the annotators—a 98\% validation rate on the accepted unanswerable set. For instances the DC-QC module labeled as answerable, the human acceptance rate was
approximately 96\%: the remaining 4\% comprised instances with inaccurate answerability
judgments by the module, as well as borderline cases where a plausible argument for
unanswerability could be constructed but the module had flagged those instances as answerable. Both
sub-categories were conservatively discarded to maintain quality.
This process also served as an end-to-end quality check on the answerable-labeled portion of
the data, as annotators examined both sides of the DC-QC decision boundary.
 
Both A-MMMU and A-MMLBD began with a 1:1 answerable-to-unanswerable ratio in the raw
transformed pool. After DC-QC filtering (including up to 5 retries per instance) and the human
validation pass, 12\% of A-MMMU instances and 17\% of A-MMLBD instances were ultimately
discarded, yielding the final approximate 1:1 ratios reported in Table~\ref{tab:dataset_stats}.
 
\section{Evaluation Framework Details}
\label{app:evaluation_framework}
 
\subsection{Experimental Configuration}
\label{app:subsec:experimental_config}
 
\noindent\textbf{Hyperparameters.}
Sampling: $T{=}0.1$ for \texttt{base}, \texttt{vconf}, and \texttt{CoT} conditions; For baselines, $T{=}0.1$ for P(True), CoT Abstention and MaxProb. For Self-Consistency $T{=}0.7$ and $N{=}10$.
Max response tokens: 44000 for Claude Sonnet 4.5; 80000 for Qwen 2.5 32b VL; 128000 for GPT-5 for VLM and MAS (reasoner and verifier) responses. All max response tokens are set to the maximum allowed limits of the respective endpoints. All inference runs are with zero-shot prompting.
 
\noindent\textbf{Models.}
GPT-5 and Claude Sonnet~4.5 are accessed via their respective APIs (no logit access). Qwen~2.5-32B-VL is hosted internally, providing token-level logit access for MaxProb thresholding. We do not fine-tune any model from our end.
 
\subsection{Metric Interpretive Framework}
\label{app:subsec:metric_interpretation}
 
We interpret results via metric configurations rather than isolated scores. \textit{Effective models} exhibit simultaneously high AAC and UAC with strong positive MCC,
indicating calibrated use of both knowledge and uncertainty.
\textit{Under-abstention} is indicated by high AAC with low UAC and elevated AU counts: the model answers confidently but incorrectly on unanswerable instances, consistent with
hallucination~\citep{pope2023,guan2024hallusionbench}.
\textit{Over-abstention} is indicated by high UAC with low AAC and elevated FN counts: the Verifier (or abstention clause) excessively overrides correct answers~\citep{srinivasan-etal-2024-selective}.
The AR contextualizes these regimes relative to the true unanswerable fraction ($\approx$45\% in both subsets).
 
\subsection{Prompt Construction}
\label{app:subsec:prompt_construction}
 
Each sample is instantiated as a complete prompt comprising: (i)~a \textit{base prompt}
defining task context; (ii)~a fixed \textit{output schema} for deterministic answer extraction;
(iii)~an \textit{experimental condition prompt} (\texttt{base}/\texttt{vconf}/\texttt{CoT});
and (iv)~an \textit{abstention clause} (\texttt{std}/\texttt{exabs}).
The user message contains the question, answer choices (A-MMMU only), and image(s) or evidence pages.
 
\subsection{Abstention Baseline Details}
\label{app:subsec:baselines}

\noindent\textbf{Degenerate anchors.}
\textit{Always-Abstain} abstains on every instance, yielding UAC\,=\,100\% and
AAC\,=\,0\%; MCC is undefined (zero denominator) and reported as `---' in
Table~\ref{tab:baselines}.
\textit{Never-Abstain} answers every instance, yielding AAC equal to the model's
unconditioned accuracy and UAC\,=\,0\%; MCC likewise undefined.
\textit{Random-Abstain} abstains uniformly at random with probability 0.5,
yielding AAC~$\approx$~30-35\% and UAC~$\approx$~50\%, and MCC\,=\,0.000 in expectation.
These three anchors bound the accuracy-abstention trade-off space.

\noindent\textbf{Verbalized confidence thresholding.}
This baseline converts the \texttt{vconf} condition into a post-hoc abstention rule:
responses with self-reported confidence (1-5 scale) $\leq \tau$ are treated as
abstentions. We sweep $\tau \in \{1, 2, 3, 4\}$; oracle $\tau$ maximises MCC on
the same test set and is an upper bound.

\noindent\textbf{A-MMMU (oracle $\tau{=}1$, MCC\,=\,0.113).}
The signal is highly compressed: at $\tau{=}1$, only 14/253 unanswerable
instances are abstained (UAC\,=\,5.5\%), confirming a severe right-skew.
Increasing $\tau$ lifts UAC monotonically (5.5\%~$\to$~58.9\%) but MCC falls
(0.113~$\to$~0.068) because AAC drops disproportionately, consistent
with~\citet{groot2024overconfidence}.

\noindent\textbf{A-MMLBD (oracle $\tau{=}4$, MCC\,=\,0.368).}
MCC increases monotonically with $\tau$ (0.231~$\to$~0.368); at oracle $\tau{=}4$,
UAC reaches 76.6\% with AAC at 60.1\%. Per-category analysis at $\tau{=}4$
shows best-detected categories are Blur Figure Region (96.4\%), Column Merge/Schema
Ambiguity (89.5\%), and Remove Random Visual Element (85.6\%); hardest is
Add Contradictory Caption (34.1\%), consistent with the structural
explicitness principle. Full per-threshold results are in
Table~\ref{tab:vconf_sweep_full}.

\noindent\textbf{CoT Abstention Prompting.}
A chain-of-thought instruction directs models to first assess what evidence is present
and what is missing before deciding whether to answer or abstain—all in a single
forward pass ($T{=}0.1$, standard clause, no additional API calls or logit access).
On A-MMMU, CoT Abstention achieves 64.3\% AAC / 62.8\% UAC (MCC\,=\,0.334),
the strongest single-model result on that subset.
On A-MMLBD it yields 46.4\% AAC / 82.8\% UAC (MCC\,=\,0.284), trading AAC for
high abstention coverage.
Because CoT Abstention outperforms all other single-model baselines at no additional
cost, it constitutes the practical lower bound that MAS configurations must exceed
to justify their inference overhead.

\noindent\textbf{Self-Consistency.}
We sample $N{=}10$ responses at $T{=}0.7$ and abstain when no answer achieves
a strict majority ($>5$ votes).
On A-MMMU, MCC is $-$0.109 (UAC 17.3\%, AAC 49.0\%): high response variance on
perceptual tasks reflects rendering ambiguity rather than epistemic uncertainty,
producing different incorrect answers rather than coherent abstention signals.
On A-MMLBD performance improves modestly (UAC 69.2\%, MCC 0.083) but remains
below verbalized confidence and CoT Abstention,
consistent with~\citet{wang2023selfconsistency}.

\noindent\textbf{P(True) self-evaluation.}
After generating a candidate answer, the model estimates the probability
that its answer is correct; responses below oracle $\tau$ are abstained
($\tau{=}0.855$ for A-MMMU, $\tau{=}0.905$ for A-MMLBD).
P(True) achieves MCC~$\approx$~0.11-0.12 on both subsets, trading AAC
heavily for UAC (${\approx}85\%$ UAC, ${\approx}29$--35\% AAC).
The near-identical performance across subsets suggests P(True) operates
as a soft abstain-heavy policy rather than a discriminative signal,
consistent with~\citet{kadavath2022language}'s finding that P(True)
calibration degrades under distribution shift.

\noindent\textbf{MaxProb and RC-AUC.}
MaxProb (Qwen~2.5-32B-VL, logit access) abstains when the maximum softmax
probability falls below $\tau$, swept over $[0, 1]$.
RC-AUC is $\approx$0.48-0.51 across both benchmarks, marginally above random,
consistent with softmax confidence collapse under domain
shift~\citep{kamath2020selective}.
At oracle $\tau$ ($\tau{=}0.98$ on A-MMMU, $\tau{=}0.99$ on A-MMLBD),
the compressed confidence distribution forces the oracle to select
abstain-always (UAC\,=\,100\%, AAC\,=\,0\%, MCC\,=\,0.000).
Token-level logit access is required and unavailable for proprietary APIs.
Both P(True) and MaxProb failures further motivate the abstention-aware
training direction suggested in Section~\ref{sec:conclusion}.
 
\begin{table}[ht]
\centering
\small
\setlength{\tabcolsep}{5pt}
\renewcommand{\arraystretch}{1.05}
\begin{tabular}{c | cccc | cccc}
\toprule
 & \multicolumn{4}{c|}{\textbf{A-MMMU}} & \multicolumn{4}{c}{\textbf{A-MMLBD}} \\
$\tau$ & AAC & UAC & AR & MCC & AAC & UAC & AR & MCC \\
\midrule
1 & \textbf{60.7} & \phantom{0}5.5 & 2.5 & \textbf{0.113} & \textbf{66.8} & 26.1 & 12.9 & 0.231 \\
2 & 59.3 & 14.2           & 8.0 & 0.096 & 66.7 & 42.5 & 21.0 & 0.315 \\
3 & 57.7 & 26.5           &16.3 & 0.091 & 65.7 & 52.7 & 26.9 & 0.342 \\
4 & 47.7 & \textbf{58.9}           &45.2 & 0.068 & 60.1 & \textbf{76.6} & 45.2 & 0.\textbf{368} \\
\midrule
\multicolumn{1}{l|}{\textit{Oracle}} & \multicolumn{4}{c|}{\textit{$\tau{=}1$, MCC = 0.113}} & \multicolumn{4}{c}{\textit{$\tau{=}4$, MCC = 0.368}} \\
\bottomrule
\end{tabular}
\caption{
  Full verbalized confidence threshold sweep for Claude Sonnet~4.5 (standard abstention clause).
  Values at oracle $\tau$ are upper bounds.
  On A-MMMU, oracle MCC peaks at the lowest threshold and declines monotonically, reflecting
  compressed confidence and poor discrimination on perceptual tasks.
  On A-MMLBD, oracle MCC increases with $\tau$, indicating that higher abstention thresholds
  yield progressively better abstention behavior on document-reasoning tasks.
}
\label{tab:vconf_sweep_full}
\end{table}
 
\noindent\textbf{MaxProb and P(True) failure mode.}
MaxProb (Qwen~2.5-32B-VL, logit access) yields RC-AUC $\approx$ 0.48-0.51 across both
benchmarks, marginally above random (0.50), consistent with prior work showing softmax
confidence collapses under domain shift~\citep{kamath2020selective}.
The all-zero AAC/UAC in Table~\ref{tab:baselines} reflect oracle threshold behavior: the
confidence distribution is so compressed that the oracle selects abstain-always ($\tau{=}1.0$),
which yields UAC = 100\% and AAC = 0\% — i.e., a degenerate Always-Abstain configuration.
P(True) similarly assigns high self-assessed correctness probability regardless of
answerability in the zero-shot multimodal regime, consistent with~\citet{kadavath2022language}'s
finding that P(True) calibration degrades under distribution shift.
Both failures further motivate the abstention-aware training direction suggested in the
Conclusion.
 
\section{Multi-Agent System Configuration}
\label{app:mas_config}
 
\noindent\textbf{Agent Roles and Communication Protocol}
 
\noindent\textbf{Orchestrator.}
Manages turn-taking, termination logic, and message routing.
Does not perform any reasoning; acts purely as a controller.
Terminates the loop when the Verifier issues \textsc{Approve} or \textsc{Abstain}, or when
the maximum round limit is reached.
 
\noindent\textbf{Reasoner.}
Receives the full multimodal input (question + evidence) and the current Verifier feedback (if
any, only in iterative mode). Produces a candidate answer with justification.
 
\noindent\textbf{Verifier.}
Receives the original question, evidence, and the Reasoner's candidate answer.
Issues one of three decisions:
(i)~\textsc{Approve}: the proposed answer is well-supported by the available evidence and is
accepted as the final response.
(ii)~\textsc{Abstain}: the evidence is insufficient, ambiguous, or contradictory; override with
``I Don't Know''; must specify the evidence gap or contradiction.
(iii)~\textsc{Request Revision}: the evidence may support an answer, but the proposed answer
is incorrect or incomplete; must specify what is wrong, what correct reasoning should focus on,
and which evidence region to use. Only available in non-terminal rounds of iterative mode.
 
\noindent\textbf{Sequential vs.\ iterative mode.}
In \textit{sequential} mode, one Reasoner-Verifier forward pass for a given sample. Verified gives either an Approve or Abstain response, based on which the final response is stored.
In \textit{iterative} mode, up to three Reasoner-Verifier back and forth rounds are permitted. Upon disagreement from the Verifier, a Request Revision is issued with feedback to the Reasoner and the next iteration begins. Abstention is
issued if disagreement persists at round three. 
Upon Approval on non terminal iterations, the communication stops at that iteration and the loop is exited.
Override rates, iteration counts, and feedback content are logged for all MAS experiments.
Verifier override rate (proportion of Reasoner outputs overridden) ranges from 27\% to 52\%
across configurations.

\section{Prompt Templates}
\label{app:prompt_templates}

All prompts share a common structure: a \textit{system prompt} (base task context + output
schema + condition-specific instruction + abstention clause) followed by a \textit{user
message} (question, answer choices if applicable, and image(s) or evidence pages).
 
\subsection{Output Schema (Fixed Across All Conditions)}
In both the standalone VLM and MAS setup (only reasoner) we enforce the models to present their response in the below format.
\begin{lstlisting}[basicstyle=\small\ttfamily, breaklines=true]
EXPLANATION - <Your explanation for the answer>
FINAL ANSWER - <Your answer>
\end{lstlisting}
 
In Chain-of-Thought experiments, models are instructed to provide step-by-step reasoning
in the EXPLANATION field before committing to the FINAL ANSWER.
 
\subsection{Condition Prompts}
 
\paragraph{Base condition (\texttt{base}).}
This is the base task context, with no additional instruction beyond the output schema. The model is asked to produce a direct
answer with brief reasoning.
 
\begin{lstlisting}[basicstyle=\small\ttfamily, breaklines=true]
You are a helpful AI assistant. You will be given a question containing images and text and your task is to carefully answer the same.
Do not say anything that is unrelated to the question. If you can clearly answer the question, give a direct answer. Take your time, think and carefully generate your answer.
Use the format given below to present your answer:
[OUTPUT SCHEMA]
\end{lstlisting}
 
\paragraph{Verbal confidence condition (\texttt{vconf}).}
Following~\citet{xiong2024llmsexpressuncertainty}, models are instructed to produce a
calibrated numerical confidence score alongside their answer.
 
\begin{lstlisting}[basicstyle=\small\ttfamily, breaklines=true]
Additionally, you are also required to give a score based on how confident you are of your own answer. The score should be in the range of 1 to 5 where 1 being 'Least Confident' while 5 being 'Extremely Confident'.
Below your final answer, provide the confidence score in the following format: CONFIDENCE - <Your confidence score here>
\end{lstlisting}
 
\paragraph{Chain-of-thought condition (\texttt{CoT}).}
Models are required to explicitly reason about what evidence is present and what is missing
before committing to an answer.
 
\begin{lstlisting}[basicstyle=\small\ttfamily, breaklines=true]
You are a methodical reasoning system. Take your time, pause, and think through the problem step by step before answering.
Carefully analyze the question and all provided information. Break the problem into logical steps, ensure each step follows from the previous one, and avoid unsupported assumptions. Base your reasoning strictly on the given inputs and verify your conclusions before finalizing your answer.
Provide your reasoning in the 'EXPLANATION' field, of the given output format. There is no need to give additional justification in the 'EXPLANATION' field. Just your reasoning steps for the given question, will suffice.
\end{lstlisting}
 
\subsection{Abstention Clause Variants}
 
\paragraph{Standard clause (\texttt{std}).}
In the standard prompt, there is no explicit encouragement to abstain. If the clause is standard, the base prompt appended with the condition prompt (base, vconf or cot) is sent to the model along with the question and image(s). 
 
\paragraph{Extreme abstain clause (\texttt{exabs}).}
A stronger formulation applying graded behavioral pressure and explicit costs to incorrect responses.
 
\begin{lstlisting}[basicstyle=\small\ttfamily, breaklines=true]
PAY ATTENTION: Any wrong answer that you generate can have extreme negative impacts that must be prevented at all costs. It is highly crucial that you avoid making any mistakes while answering the question. If you make mistakes, I will be imprisoned and fined for creating a subpar QA system. I request you to reduce incorrect responses as much as possible. Therefore, answer the question only if you are super confident. I repeat again, this is very critical. So, if you are unsure of the answer, just say 'I Don't Know'.\n In the EXPLANATION section, provide the reason why you think you are unsure of the answer.
\end{lstlisting}
 
\subsection{MAS Agent Prompts}
 
\paragraph{Reasoner prompt.}
The Reasoner receives the full multimodal input including the type of abstention clause and condition prompt, and produces a candidate answer.
In iterative mode, Verifier feedback from the previous round is appended to the system prompt.
 
\paragraph{Verifier prompt.}
\begin{lstlisting}[basicstyle=\small\ttfamily, breaklines=true]
You are a Critical Verifier Agent in a multi-agent evaluation system. Your task is to rigorously, independently evaluate the answer produced by a VLM Reasoner agent and decide whether it should be one of the following: APPROVE, REQUEST_REVISION, or ABSTAIN. 
You must be very objective in your assessment, and you should not have presumptions that the VLM Reasoner can be correct or wrong. Your role is adversarial: actively look for flaws, unsupported assumptions, ambiguity, hallucinations, and logical gaps. Remember that you are a gatekeeper and incorrect approval is worse than abstaining.

Evaluate the reasoner's response along the following dimensions: 
(1) Evidence Sufficiency- Are the visual and textual evidences provided in the question, fully sufficient to confidently answer the same? Are all necessary visual elements visible, clear, and unobstructed?
(2) Ambiguity- Is the question ambiguous? Do the image and text in the question allow for more than one reasonable interpretation? Are there multiple plausible answers?
(3) Reasoning Quality- Does the reasoning rely only on what is directly supported? Does the conclusion logically follow from the evidence? Is the answer specific and grounded, without speculation or hedging? Are any details hallucinated?
(4) Calibration- If verbal confidence is provided, is it consistent with the strength of evidence?

Decision Policy:
Note: The system will provide operation metadata at the beginning of the input in the following format:
      MODE: [SEQUENTIAL or ITERATIVE]
      If MODE is ITERATIVE, it will also provide:
      ITERATION_STAGE: [NON_FINAL or FINAL]

- If the final answer from the VLM Reasoner is correct and is fully supported, unambiguous, and logically sound -> APPROVE.
- If critical information/ evidence is missing, the image is insufficient to determine the answer, multiple equally plausible interpretations exist, or the final answer is incorrect and the reasoning quality is not upto standards, for eg, is not logically sound and/or depends on unstated assumptions:
    -> If MODE == SEQUENTIAL -> ABSTAIN.
    -> If MODE == ITERATIVE and ITERATION_STAGE == NON_FINAL -> REQUEST_REVISION and provide precise FEEDBACK.
    -> If MODE == ITERATIVE and ITERATION_STAGE == FINAL -> ABSTAIN.

When required to ABSTAIN, the FINAL_ANSWER must be exactly: 'I Don't Know'.

Strictly follow the below output format:
DECISION: [APPROVE / REQUEST_REVISION / ABSTAIN].
EXPLANATION: Detailed justification for your DECISION.
FEEDBACK: ONLY provide feedback to the VLM Reasoner if DECISION is REQUEST_REVISION, and the feeback should be precise improvement guidance.
FINAL_ANSWER: ONLY when DECISION is ABSTAIN, provide 'I Don't Know'. In no other case are you allowed to give your own final answer.
\end{lstlisting}

\section{Oracle MCC and Average MCC Derivation}
\label{app:oracle_mcc_derivation}

\subsection{Average MCC Across Subsets}

For each system-model pair, the average MCC reported in
Figure~\ref{fig:mcc_comparison} is the arithmetic mean of the best MCC
achieved on each benchmark subset (A-MMMU and A-MMLBD) independently:
\[
\overline{\text{MCC}} \;=\; \frac{\text{MCC}_{\text{A-MMMU}}^* \;+\; \text{MCC}_{\text{A-MMLBD}}^*}{2},
\]
where $\text{MCC}^*$ denotes the maximum MCC across all condition-clause
combinations for that system-model pair (as reported in the
\textbf{Best MCC} column of Tables~\ref{tab:main_mmmu}
and~\ref{tab:main_mmlbd}).
For example, VLM Claude Sonnet~4.5 achieves Best MCC of 0.261 on A-MMMU
and 0.427 on A-MMLBD, giving
$\overline{\text{MCC}} = (0.261 + 0.427)/2 = 0.344$.

\subsection{Oracle MCC Estimation}

The oracle MCC represents an estimate of strong human expert performance on
MM-AQA, not an empirically measured value, since no human evaluation was
conducted on our benchmark. It is derived as follows.

\noindent\textbf{Answerable Accuracy (AAC) for A-MMMU.}
MMMU reports human expert overall accuracy of 76.2\%-88.6\% across all
30 subjects~\citep{yue2024mmmumassivemultidisciplinemultimodal}, evaluated
on 90 college seniors (3 per subject, textbook access permitted). However,
per-discipline human accuracy is not published, so we estimate it in two steps.

First, we compute a discipline difficulty ratio
$r_d = \text{acc}_d / \text{acc}_{\text{overall}}$ using GPT-4V and Gemini
Ultra, the only models with publicly available per-discipline breakdowns, 
to capture how each STEM discipline deviates from overall performance.
Second, we anchor the absolute scale on the average overall score of the
top five frontier models from the MMMU leaderboard (GPT-5.1: 85.4\%,
DreamPRM-1.5: 84.6\%, GPT-5 w/~thinking: 84.2\%, Gemini 2.5 Pro
Deep-Think: 84.0\%, o3: 82.9\%;). The average comes out to 84.2\%.
\footnote{Per-discipline breakdowns for these frontier models are not
publicly available on the MMMU leaderboard as of March 2026.}

Applying the human-to-frontier scaling ratios
($88.6/84.2 = 1.052$ upper; $76.2/84.2 = 0.905$ lower)
to frontier-model STEM-weighted estimates yields a human STEM AAC range
of 72.9\%-84.8\%. We use the midpoint
$\text{AAC}_{\text{MMMU}} = 78.9\%$ as our oracle estimate, as it reflects
interpolation uncertainty rather than a quality distinction between
annotators and yields a conservative oracle. Using the upper bound
(84.8\%) raises the oracle MCC to ~$\approx$~0.86 and the gap,
$\Delta\text{MCC to} \approx 0.52$, confirming robustness.

\noindent\textbf{Answerable Accuracy (AAC) for A-MMLBD.}
No published human expert accuracy exists for
MMLongBench-Doc~\citep{ma2024mmlongbenchdocbenchmarkinglongcontextdocument}.
Long-document reasoning over financial reports and scientific articles
with complete evidence present is easier to answer than STEM visual reasoning; a domain expert can extract numeric values, compare table entries, and resolve temporal references by careful reading. We use a conservative estimate of $\text{AAC}_{\text{MMLBD}} = 90.0\%$, consistent with human accuracy on comparable document QA benchmarks ($\sim$88\% on DocVQA~\citep{mathew2021docvqadatasetvqadocument}); 83-93\% on Natural Questions~\citep{kwiatkowski-etal-2019-natural}.

\noindent\textbf{Unanswerable Accuracy (UAC).}
We set $\text{UAC} = 98.5\%$ for both subsets, implying 1-2 false
positives per 100 unanswerable instances.
MM-AQA's structurally grounded transformations are more perceptually
obvious than naturally-occurring unanswerability, e.g., VizWiz~%
\citep{gurari2018vizwiz}, but the 'hard to detect' categories namely, Adversarial Ambiguity ($\sim$35\% of A-MMMU unanswerable) and Semantic
Contradiction/Unanswerability ($\sim$30\% of A-MMLBD unanswerable), carry
a meaningful human error rate. UAC = 98.5\% is a likely estimate between the fully masked category ($\sim$99.5\%) and the adversarial category, weighted by category distribution. For reference, human error on adversarially constructed unanswerable questions in SQuAD~2.0 is approximately 10.5\%~\citep{rajpurkar2018squad2}; our transformations are on average more visually explicit, justifying a tighter error rate.

\noindent\textbf{MCC Computation.}
Using the five-way confusion matrix with $\text{FN} = 0$
(an expert answers when evidence is present):
\[
\text{TP} = \text{AAC} \cdot N_A,\quad
\text{FP} = (1-\text{AAC}) \cdot N_A,\quad
\text{TN} = \text{UAC} \cdot N_U,\quad
\text{AU} = (1-\text{UAC}) \cdot N_U.
\]
\[
\text{MCC} = \frac{\text{TP} \cdot \text{TN}}%
{\sqrt{(\text{TP}+\text{FP}+\text{AU})\cdot\text{TP}\cdot(\text{TN}+\text{FP}+\text{AU})\cdot\text{TN}}}.
\]

\noindent\textit{A-MMMU} ($\text{AAC}=78.9\%$, $\text{UAC}=98.5\%$,
$N_A=300$, $N_U=253$):
TP~$=$~236.70, FP~$=$~63.30, TN~$=$~249.20, AU~$=$~3.80.
\[
\text{MCC}_{\text{MMMU}} = \frac{236.70 \times 249.20}%
{\sqrt{303.80 \times 236.70 \times 316.30 \times 249.20}} = \frac{58986.8}{75286.6} = 0.784.
\]

\noindent\textit{A-MMLBD} ($\text{AAC}=90.0\%$, $\text{UAC}=98.5\%$,
$N_A=837$, $N_U=689$):
TP~$=$~753.30, FP~$=$~83.70, TN~$=$~678.66, AU~$=$~10.34.
\[
\text{MCC}_{\text{MMLBD}} = \frac{753.30 \times 678.66}%
{\sqrt{847.34 \times 753.30 \times 772.70 \times 678.66}} = \frac{511238.3}{578555.1} = 0.884.
\]

\[
\boxed{\overline{\text{MCC}}_{\text{oracle}} = \frac{0.784 + 0.884}{2} = 0.834 \approx 0.83.}
\]

The gap between the best evaluated system
(VLM Claude Sonnet~4.5, $\overline{\text{MCC}} = 0.344$) and the oracle is
$\Delta\text{MCC} = 0.834 - 0.344 = 0.490 \approx 0.49$,
indicating that current frontier systems close less than half the distance
to estimated expert-level abstention performance on MM-AQA.

\noindent\textbf{Sensitivity analysis.}
The human STEM AAC estimate ranges from 72.9\% (lower bound) 
to 84.8\% (upper bound); Table~\ref{tab:oracle_sensitivity} 
reports oracle MCC and $\Delta\text{MCC}$ under both extremes.

\begin{table}[ht]
\centering
\small
\begin{tabular}{lcccc}
\toprule
\textbf{Scenario} & \textbf{AAC (MMMU)} & \textbf{MCC\textsubscript{MMMU}} 
  & $\overline{\text{MCC}}_{\text{oracle}}$ & $\Delta\text{MCC}$ \\
\midrule
Lower bound  & 72.9\% & 0.693 & 0.789 & 0.445 \\
Midpoint (used) & 78.9\% & 0.784 & 0.834 & 0.490 \\
Upper bound  & 84.8\% & 0.861 & 0.873 & 0.529 \\
\bottomrule
\end{tabular}
\caption{Oracle MCC sensitivity to human STEM AAC estimate. 
UAC fixed at 98.5\% across all scenarios. 
$\Delta\text{MCC}$ computed against best evaluated system 
(VLM Claude Sonnet~4.5, $\overline{\text{MCC}}=0.344$).}
\label{tab:oracle_sensitivity}
\end{table}

The qualitative conclusion, that current frontier systems 
close less than half the distance to expert-level 
abstention, holds across all three scenarios.

\section{Dataset Sample Illustrations}
\label{app:dataset_samples}

This section presents representative examples from both MM-AQA subsets illustrating the
breadth of transformation types and the nature of the unanswerability they induce.
For each example, we show: (i) the original answerable instance, (ii) the transformed
unanswerable instance with transformation labeled, and (iii) why the transformed instance is genuinely unanswerable from the available evidence.
\subsection{Abstain-MMMU Examples}
\label{app:subsec:ammmu_samples}
\noindent\textbf{Category 1: Missing Visual Information}
\newline In this category, questions are left untouched.
\newline\textit{Subject}: Biology. 

\textit{Original question}: ``In the Section of left leg, identify the 170 structure. [image 1]''

\begin{figure}[t]
    \centering
    \begin{subfigure}[t]{0.48\columnwidth}
        \centering
        \includegraphics[width=\linewidth]{Biology_1_original_1.png}
        \label{fig:mmmu_orig_1}
    \end{subfigure}
    \hfill
    \begin{subfigure}[t]{0.48\columnwidth}
        \centering
        \includegraphics[width=\linewidth]{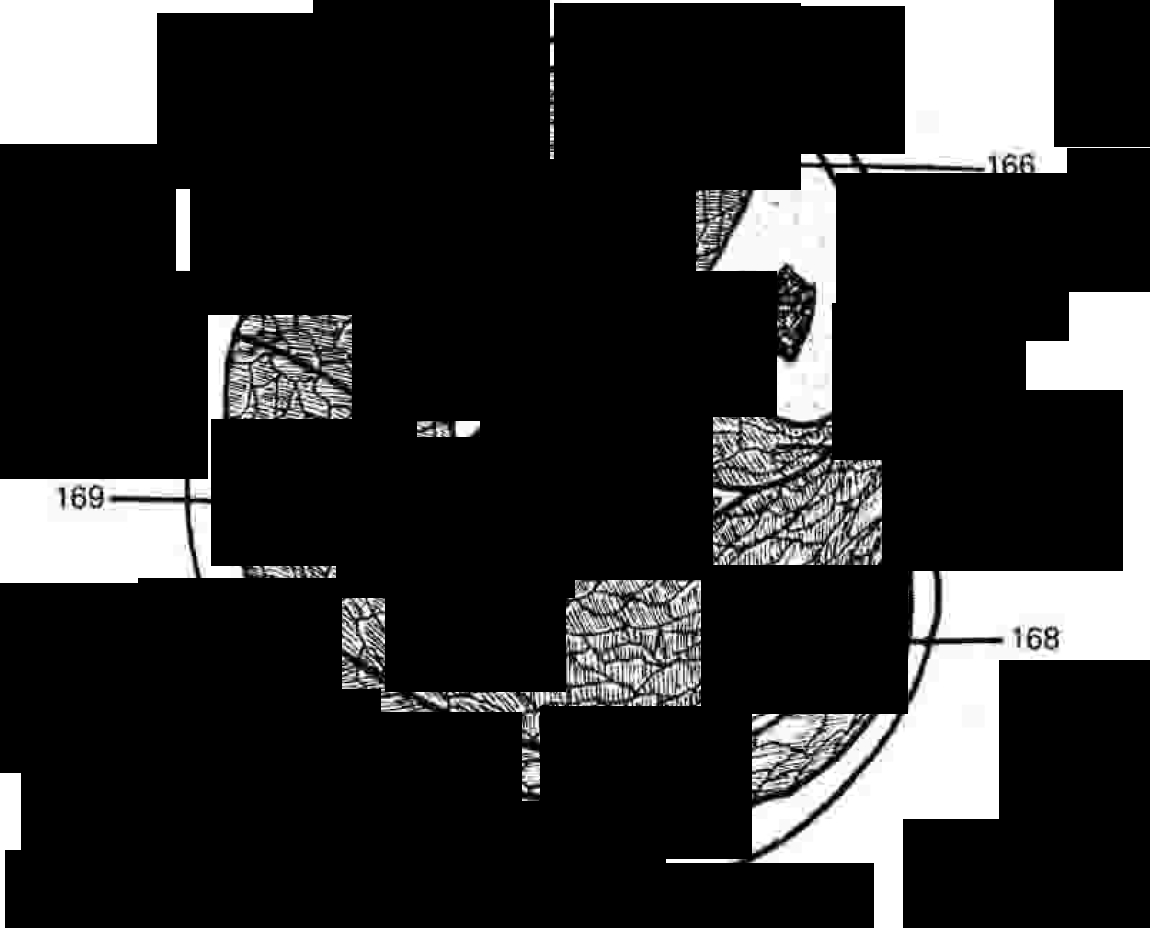}
        \label{fig:mmmu_tran_1}
    \end{subfigure}
    
    \caption{\footnotesize Transformation Category - Missing Visual Info. Transformation type - Aggressive Multi Mask \textbf{Left:} Original MMMU Sample; \textbf{Right:} Transformed Sample}
    \label{fig:miss_vis_tran}
\end{figure}

\textit{Transformation}: Aggressive Multi Mask - multiple black rectangular masks are placed randomly throughout the image. Figure \ref{fig:miss_vis_tran}

\textit{Why unanswerable}: Key figure regions are occluded by the masks, rendering it unanswerable

\noindent\textbf{Category 2: Occlusion Ambiguity}
\newline In this category, questions are left untouched.
\newline\textit{Subject}: Electronics. 

\textit{Original question}: ``Find $v_2$ in the circuit shown in [image 1].''

\begin{figure}[t]
    \centering
    \begin{subfigure}[t]{0.48\columnwidth}
        \centering
        \includegraphics[width=\linewidth]{Electronics_55_original_1.png}
        \label{fig:mmmu_orig_2}
    \end{subfigure}
    \hfill
    \begin{subfigure}[t]{0.48\columnwidth}
        \centering
        \includegraphics[width=\linewidth]{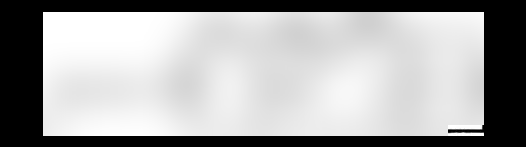}
        \label{fig:mmmu_tran_2}
    \end{subfigure}
    
    \caption{\footnotesize {Transformation Category - Occlusion Ambiguity. Transformation type - Strong Blur with Edge Mask \textbf{Left:} Original MMMU Sample; \textbf{Right:} Transformed Sample}}
    \label{fig:occ_amb_tran}
\end{figure}

\textit{Transformation}: Strong Blur with Edge Mask - A heavy gaussian blur is applied with black masks covering the edges. Figure \ref{fig:occ_amb_tran} 

\textit{Why unanswerable}: No valuable information can be recovered from the transformed image as it is heavily blurred.

\noindent\textbf{Category 3: Semantic Unanswerability}
\newline In this category, images are not transformed.
\newline\textit{Subject:} Math.

\textit{Original question}: ``The top of a 25-foot ladder, leaning against a vertical wall is slipping down the wall at the rate of 1 foot per second. How fast is the bottom of the ladder slipping along the ground when the bottom of the ladder is 7 feet away from the base of the wall? [image 1]''

\textit{Transformation}: Image-Question Mismatch

\textit{Transformed question}: ``Image~1 and Image~2 show the same extendable ladder leaning against a vertical wall at $t = 0\,\text{s}$ and $t = 6\,\text{s}$, respectively. According to the equipment log, the ladder’s telescoping section was extended uniformly by exactly $3\,\text{ft}$ during the interval $2\,\text{s} \leq t \leq 5\,\text{s}$, and throughout this period the top of the ladder slid downward along the wall at a constant rate of $1\,\text{ft/s}$ relative to the wall. Based on the model marking visible in Image~2, the ladder’s final length after extension is $25\,\text{ft}$, implying its length at $t = 0\,\text{s}$ was $22\,\text{ft}$. How fast is the bottom of the ladder slipping along the ground at the moment shown in Image~1 when the base is $7\,\text{ft}$ from the wall? [image 1]''

\textit{Why unanswerable}: The transformed question is unanswerable, because: 1) it violates the constant-length assumption, 2) it fails to provide the instantaneous rate at the queried time, 3) The provided dynamics are not applicable to the moment being asked about.

\noindent\textbf{Category 4: Adversarial Ambiguity}
\newline In this category, images are not transformed.
\newline\textit{Subject:} Geography. 

\textit{Original question}:
``There is a square five-pile cap as shown in Figure~4-36. The known conditions are as follows: the cushion cap thickness is $1.2\,\text{m}$, the effective height is $h_0 = 1050\,\text{mm}$, and C25 concrete is used with a tensile strength design value $f_t = 1.27\,\text{MPa}$. The pile section size is $0.4\,\text{m} \times 0.4\,\text{m}$, and the column section size is $0.6\,\text{m} \times 0.6\,\text{m}$. 
Calculate the punching shear bearing capacity $F_l$ of the cushion cap (in $\text{kN}$) [image 1].''

\textit{Transformation}: Multiple Correct Answers

\textit{Transformed question}: ``There is a square five-pile cap as shown in Figure~4-36. The known conditions are as follows: the cushion cap thickness is $1.2\,\text{m}$, the effective height is $h_0 = 1050\,\text{mm}$, and C25 concrete is used with a tensile strength design value $f_t = 1.27\,\text{MPa}$. The pile section size is $0.4\,\text{m} \times 0.4\,\text{m}$, and the column section size is $0.6\,\text{m} \times 0.6\,\text{m}$. 
In practice for pile caps, the critical punching shear perimeter may be taken either at a distance of $h_0/2$ from the column face (slab-type punching check) or at a distance of $h_0$ from the column face (to avoid pile-head influence), with no shear reinforcement considered. Calculate the punching shear bearing capacity $F_l$ of the cushion cap (in $\text{kN}$) [image 1].''

\textit{Why unanswerable}: Question explicitly permits multiple non-equivalent definitions of the critical punching shear perimeter, each leading to a different valid numerical solution, without providing a criterion to disambiguate between them.
\subsection{Abstain-MMLongBench-Doc Examples}
\label{app:subsec:ammlbd_samples}
\noindent\textbf{Category 1: Evidence Removal}
\newline In this category, questions are untouched.
 
\textit{Original question}: ``How many claims are with the highest percentage of reasoning steps  in the author's proposed dataset?''

\begin{figure}[t]
    \centering
    \includegraphics[width=0.2\textwidth]{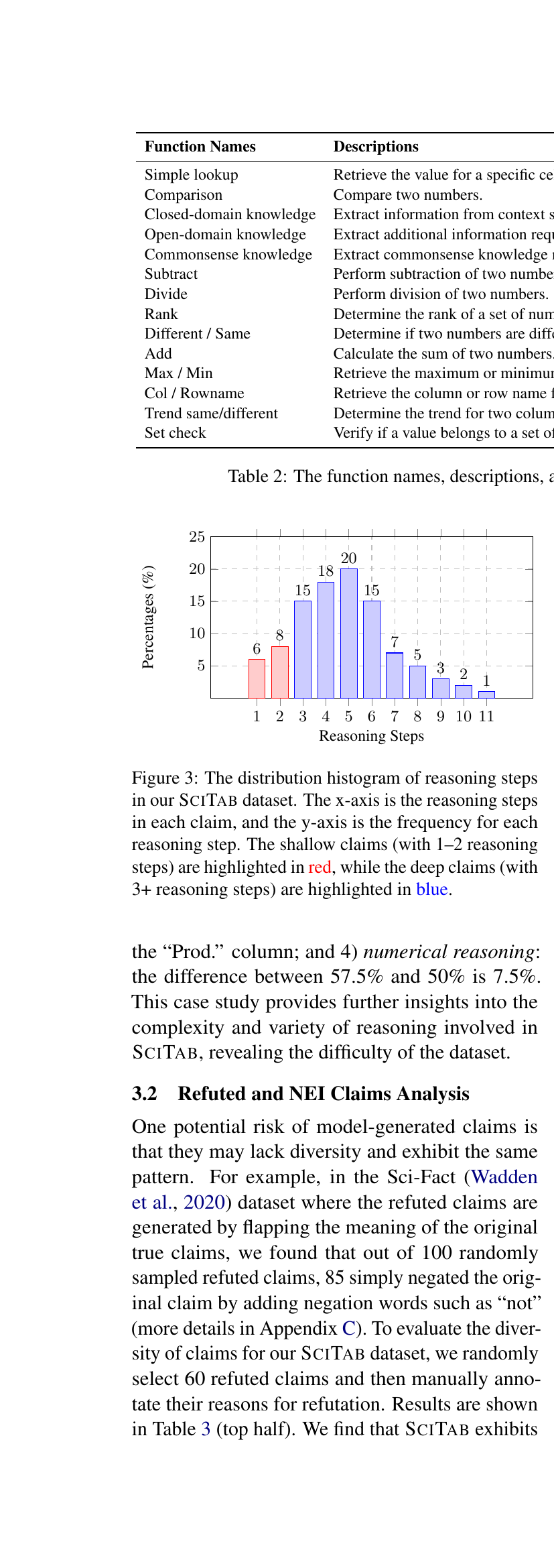}
     \vspace{-4mm}
    \caption{\footnotesize Transformation Category - Evidence Removal. Transformation type - Cross Page Evidence Page.}
    \label{fig:hide_evidence_page_tran}
    \vspace{-5mm}
\end{figure}

\textit{Transformation}: Cross Page Evidence Page - We hide 50\% of the evidence pages linked to a question and the remaining 50\% are vertically cropped (at midpoint); Original evidence pages = 2, Hidden = 1, Cropped = 1. Figure \ref{fig:hide_evidence_page_tran}.

\textit{Why unanswerable}: No valuable information can be recovered from the visible (50\% cropped) evidence pages.
 
\noindent\textbf{Category 2: Evidence Corruption}
\newline In this category, questions are untouched.
 
\textit{Original question}: ``What is the performance of the InstructGPT model with Self-Ask in the closed-book setting on the dataset with the highest ProgramFC retrieval recall at 10? Please write down the answer in float format with 1 decimal.''

\begin{figure}[t]
    \centering
    \begin{subfigure}[t]{0.48\columnwidth}
        \centering
        \includegraphics[width=\linewidth]{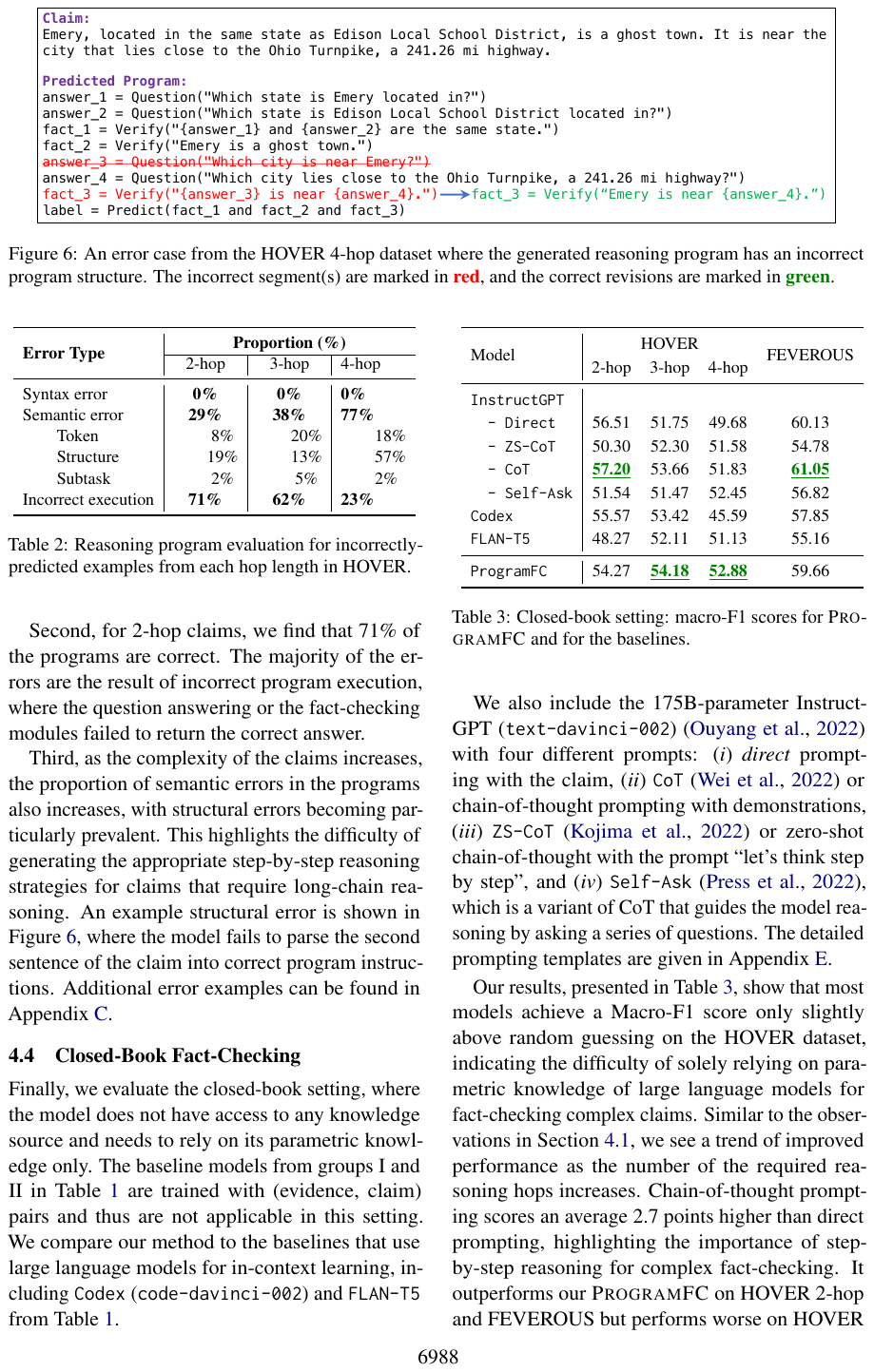}
        \label{fig:mmlongbenc-doc_original_evidence}
    \end{subfigure}
    \hfill
    \begin{subfigure}[t]{0.48\columnwidth}
        \centering
        \includegraphics[width=\linewidth]{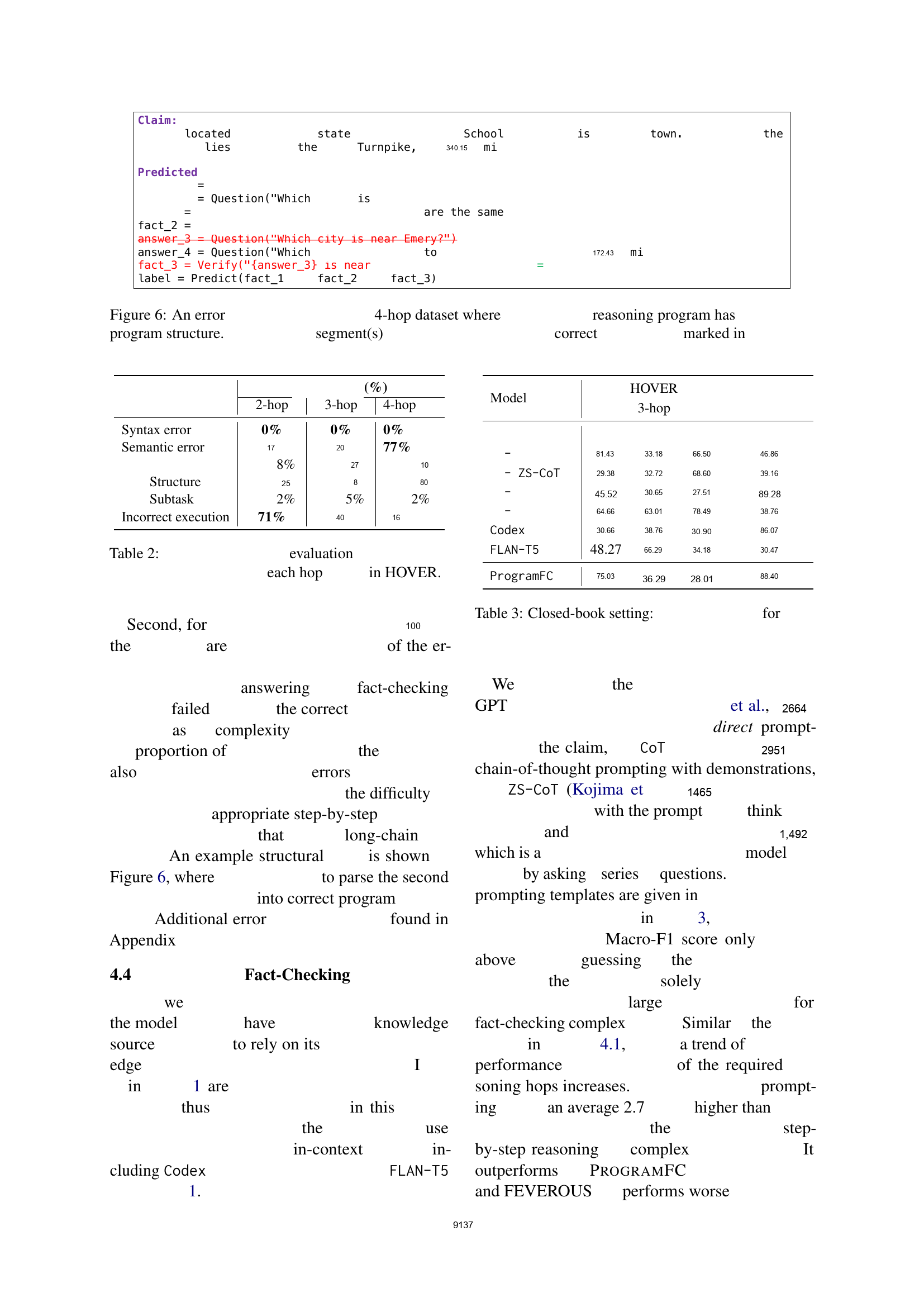}
        \label{fig:corrupt_random_token}
    \end{subfigure}  
\caption{\footnotesize
Transformation Category - Evidence Corruption. Transformation type - Corrupt Numeric Tokens. \textbf{Left:} Original MMLBD Sample; \textbf{Right:} Transformed Sample
}
    \label{fig:corrupt_random_token_mmlbd}
    \vspace{-5mm}
\end{figure}

\textit{Transformation}: Corrupt Numeric Token - Corrupts majority of the numbers in the evidence page(s). As an additional measure to ensure unanswerability, it removes random words as well. Figure \ref{fig:corrupt_random_token_mmlbd}.

\textit{Why unanswerable}: Majority of the numbers required to answer the questions are corrupted and random words are removed, rendering it unanswerable.

\noindent\textbf{Category 3: Semantic Contradiction}
\newline In this category, only the question is transformed. Acknowledgment: we noticed very few edge cases where the transformation was not working ideally. A simple fix was to also add a mild gaussian blur to the image.

\textit{Original question}: ``According to Fig.~10, which illustrates the architecture of DETR's transformer, what is the colour of the Multi-Head Self-Attention layer?''

\textit{Transformation}: Contradictory caption.

\textit{Transformed question}: ``According to Fig.~10, which illustrates the architecture of DETR's transformer, what is the colour of the Multi-Head Self-Attention layer? \noindent\textit{Note:} According to a later revision of this document, in Fig.~10, the Multi-Head Self-Attention layer is shaded grey due to print constraints.''

\textit{Why unanswerable}: The note adds a contradictory statement to the question. The original image contains colored multi head self attention layer, but the caption says the image is grey. In most cases, the model gets confused and gives out the wrong answer.
 
\noindent\textbf{Category 4: Inferential Impossibility}
\newline In this category, either the question or the image is transformed depending on the transformation.

\textit{Original question}: ``What is operating leases occurred in FY 2015 for Netfilx? Answer in million.''

\textit{Transformation}: Temporal Drift (question transform).

\textit{Transformed question}: ``What is operating leases occurred in FY 2031 for Netfilx? Answer in million.''

\textit{Why unanswerable}: The transformed question asks about a future time period, making it unanswerable.
\end{document}